\documentclass[journal]{IEEEtran}

\ifCLASSINFOpdf
\else
\fi

\usepackage{amsmath}
\usepackage{amsfonts}
\usepackage{amssymb}
\usepackage{bm}
\usepackage{mathtools, cuted}

\usepackage{rotating}
\usepackage{color}
\usepackage{cite}
\usepackage{graphicx}
\usepackage{caption}
\usepackage{subcaption}
\usepackage{threeparttable}

\usepackage{array}

\newcolumntype{C}[1]{>{\centering\let\newline\\\arraybackslash\hspace{0pt}}b{#1}}
\newcolumntype{R}[1]{>{\raggedleft\let\newline\\\arraybackslash\hspace{0pt}}b{#1}}

\begin{document}
%


\title{How Much Data is Enough? A Statistical Approach with Case Study on Longitudinal Driving Behavior}



%
%
%

\author{Wenshuo~Wang,  \IEEEmembership{Studen~Member,~IEEE,}
        Chang~Liu,  \IEEEmembership{Student~Member,~IEEE,}
        Ding~Zhao 
\thanks{W. Wang is with the Department of Mechanical Engineering, Beijing Institute of Technology, Beijing,
China 100081 and also with the Department of Mechanical Engineering, University of California at Berkeley, Berkeley, CA, 94720 USA. e-mail: wwsbit@gmail.com; wwsvdc2015@berkeley.edu.}
\thanks{C. Liu is with the Department of Mechanical Engineering, University of California at Berkeley, Berkeley, CA, 94720 USA. e-mail: changliu@berkeley.edu.}
\thanks{D. Zhao (corresponding author) is with the Department of Mechanical Engineering of the University of Michigan, Ann Arbor, MI 48109 USA. e-mail: zhaoding@umich.edu}

}

%



\maketitle

\begin{abstract}
Big data has shown its uniquely powerful ability to reveal, model, and understand driver behaviors. The amount of data affects the experiment cost and conclusions in the analysis. Insufficient data may lead to inaccurate models while excessive data waste resources. For projects that cost millions of dollars, it is critical to determine the right amount of data needed. However, how to decide the appropriate amount has not been fully studied in the realm of driver behaviors. This paper systematically investigates this issue to estimate how much naturalistic driving data (NDD) is needed for understanding driver behaviors from a statistical point of view. A general assessment method is proposed using a Gaussian kernel density estimation to catch the underlying characteristics of driver behaviors. We then apply the Kullback-Liebler divergence method to measure the similarity between density functions with differing amounts of NDD. A max-minimum approach is used to compute the appropriate amount of NDD. To validate our proposed method, we investigated the car-following case using NDD collected from the University of Michigan Safety Pilot Model Deployment (SPMD) program. We demonstrate that from a statistical perspective, the proposed approach can provide an appropriate amount of NDD capable of capturing most features of the normal car-following behavior, which is consistent with the experiment settings in many literatures.
\end{abstract}



%
\IEEEpeerreviewmaketitle

\section{Introduction}
\IEEEPARstart{N}{aturalistic} driving studies have shown great potential in smart city\cite{vilajosana2013bootstrapping,townsend2013smart}, transportation energy efficiency\cite{arias2016electric,zhou2016big,cai2014siting}, and driver behaviors \cite{klauer2006impact,green2008integrated,zhao2016accelerated}, in which  data are collected from a number of equipped vehicles driven under naturalistic conditions over an extended period of time \cite{klauer2006impact}. Research institutes around the world have spent great efforts and recourses collecting naturalistic driving data (NDD). For example, the major projects of naturalistic driving study from countries around the world such as the United States, the European Union, Australia, Japan, and China are listed in Table \ref{Major_project}. From Table \ref{Major_project}, these naturalistic driving studies vary greatly in research topics, the number of participant drivers ranging from 11 to over 2,700, and the duration of experiments ranging from 1 to 6 years. What has not been fully studied, however, is how much driving data is sufficient to address problems such as the cause of accidents, distraction and inattention, eco-driving styles, modeling driver behavior, and the effects of driver assistance systems on driver behavior. Similar problem concerning ``\textit{How much data is enough?}'' have been asked in other fields\cite{heyman2002much,wortley2005much,saris2003much,splinter2013much} such as sociology, biology, and oceanography, but not yet in the fields of analyzing/modeling human driving behaviors and traffic safety. Therefore, to avoid the issues of insufficient or excessive data and offer a guideline for primary experiment design,  we need to develop an efficient way to estimate the appropriate amount of NDD for a variety of problems. 

%
%
%
%

\begin{table*}[t] 
	\centering
	\caption{\textsc{Major  Projects of Naturalistic Driving Study in The World}}
	\begin{threeparttable}
	\begin{tabular}{C{2.3cm}|C{1.6cm}|C{1.1cm}|C{1.4cm}|C{1.2cm}|C{1.6cm}|C{2.2cm}|C{3.1cm}}
		\hline
		\hline
		Project name & Conductor & Period & Mileage [mile] & Vehicle & Sensor & Drivers & Research topic \\
		\hline
		100 Car Naturalistic Driving Study \cite{klauer2006impact} &  Virginia Tech. & 2001--2009 &  $2\times 10^{6}$ & 100 sedans & camera &109 primary drivers, 132 secondary drivers & Rear end collision \\
		\hline
		Automotive Collision Avoidance System \cite{Ervin2005impact}&  University of Michigan & 2004-2005 & $ 1.37\times 10^{5} $ & 11 sedans & camera, radar & 96 drivers & Forward collision warning (FCW) \\
		\hline
		Road Departure Crash Warning\cite{LeBlanc2006road} & University of Michigan & 2005--2006 & $ 8.3 \times 10^{4} $ & 11 sedans & camera, radar & 11 drivers & Lane departure warning (LDW)\\
		\hline 
		Sweden-Michigan Naturalistic Field Operational Test (SeMiFOT) \cite{Victor2010}& University of Michigan & 2008--2009 & $ 1.07 \times 10^{5} $ & 10 sedans, 4 trucks & camera, radar & 39 drivers & FCW, LDW, blind spot information system, electronic stability control, and impairment warning \\
		\hline
		Integrated Vehicle-Based Safety Systems\cite{sayer2011integrated} & University of Michigan & 2010--2011 &  sedans: 213\&309; trucks: 601\&944 & 16 sedans 10 heavy trucks& camera, radar & 108 drivers for sedans; 18 professional truck drivers & Integrated warning \\
		\hline 
		Safety Pilot Model Deployment \cite{bezzina2014safety} & University of Michigan & 2012--2014 &  more than $ 3.4\times 10^{7} $ & 2,800 various types of vehicles & camera, radar & 2,700 volunteer drivers and several professional bus and truck drivers & Connected vehicle\\
		\hline
		Google driverless car\cite{google} & Google & 2012--present & more than $ 1.3 \times 10^{6}$ & At least 50 sedans and SUVs & lidar, camera, radar & Google technicians and volunteers & Fully self-driven vehicle\\
		\hline
		Australian Naturalistic Driving Study or Australian 400-car Naturalistic Driving Study \cite{ANDS2015,regan2013australian} & Led by University of New South Wales & 2015--present & 4 months & 400 vehicles & camera, CAN data, GPS & 360 participants (180 in New South Wales and 180 in Victoria) & Safety at intersections; Speed choice; Interactions with vulnerable road users; Fatigue; Distraction and inattention; Crashes and near-crashes; Interactions with ITS\\
		\hline 
		European naturalistic Driving and Riding for Infrastructure \& Vehicle safety and Environment(UDRIVE) \cite{barnard2016study}& the 7th EU Framework Programme and 20 partners & 2012--2017 & On going & 200 vehicles (cars, trucks, and scooters)& cameras, IMU sensors, GPS, Mobil Eye smart camera, CAN data, and Sound level& On going &  Crash causation and risk; Everyday driving; Distraction and inattention; Vulnerable road users; Eco-driving\\
		\hline 
		China Naturalistic Driving Study & Tongji University; VTTI; General Motors&2012--2015 & more than $ 1.0 \times 10^{5} $ & 5 vehicles &  -- &  90 drivers; each drove vehicle for 2 months & Exploring Chinese moped-vehicle conflict configurations; Examining car driver responses to moped-vehicle conflicts \\
		\hline
		Japan Naturalistic Driving Study \cite{uchida2010investigation} &Ministry of Land, Infrastructure, Transport and Tourism  & 2006--2008& --& 60 vehicles (35 wagons \& 25 sedans)&GPS, CAN data, acceleration sensor, camera & 60 drivers (58 males \& 2 females)& Accident causation research\\
		\hline
		\hline
	\end{tabular}
\end{threeparttable}
	\label{Major_project}
\end{table*}

The required amount of NDD depends on the problem to be solved, the way the problem is formulated, and the dataset to be analyzed (e.g., NDD or driving simulator-based data). For example, a traffic accident analysis usually requires the data with longer driving period than that of modeling driver behaviors, because the reasons for traffic accidents are diverse and reflect a small probability event, compared to common driving behavior. Therefore, to answer this question asked by ``\textit{How much naturalistic driving data is enough in understanding driving behaviors?}'', we make a further discussion and analysis for different cases and propose a general assessment approach to determine the appropriate amount of NDD from a statistical perspective. 

In this paper, our main contributions are: (1) we introduce the problem of the amount of driving data; (2) we propose a general assessment approach to compute an appropriate amount of the required naturalistic driving data; (3) a case of modeling car-following behaviors using naturalistic driving data is conducted to validate our proposed method.

This paper is organized as follows. Section II reviews the related work and analyzes the reasons for diversity in the amount of NDD appearing in the literature. Section III presents a general assessment approach to determine the critical value for the required amount of data. Section IV presents the experiments and the results of a case study for modeling driver behavior. Section V concludes this paper with a discussion, final remarks, and future research directions.

\begin{table*}[t]
	\centering
	\caption{\textsc{The Amount of Naturalistic Driving Data in Different Studies on Modeling Driver Behaviors$^\dagger $}}
	\begin{threeparttable}
	\begin{tabular}{C{1.3cm}|C{1.5cm}|C{1.9cm}|C{1cm}|C{2.3cm}|C{3.5cm}|C{2.5cm}}
		\hline
		\hline
		References & drivers & vehicles & events & Total time $ t $ & Driving tasks & Data type\\
		\hline
		\cite{Lefevre16} & 5 & 1 & * & 300 [min]& Car following & In-vehicle sensors \\ \hline
		\cite{ossen2006interdriver} & * & 3 & 229 & $t \approx $190.8 min & Car following& Camera/video data \\ \hline 
		\cite{higgs2015segmentation} & 20 & * & 392 &  $t > $ 196 min & Car following & In-vehicle sensors\\ \hline
		\cite{qi2016appropriate} & 13 & * & * & $t > $ 1,200 min & Car following & In-vehicle sensor \\ \hline
		\cite{pariota2015linear} & * & * & 54 & $ t \approx $ 1172.8  min & Car following & In-vehicle sensors \\ \hline
		\cite{Butakov16} & 3 & * & * & * & Signalized Intersections & Camera/video data \\ \hline
		\cite{liu2016driver} & 41 & * & * & 49 $ \lesssim t \lesssim $ 184 min & Driver distraction& In-vehicle sensors \\ \hline
		\cite{li2016detecting} &*  & * & * &  $t \approx $ 720 min& Mirror-checking actions & In-vehicle sensors \\ \hline
		\cite{nobukawa2016gap}& 18 & 26 & * & * & Lane change & In-vehicle sensors \\ \hline
		\cite{Butakov15} & 3 & 1 & * &   4,947 min& Lane change  &  In-vehicle sensors \\ \hline
		\cite{zhao2017road} & * & * & $ > $ 5,700 & $ > $ 1,140 min& Lane change & Multisensor data \\ \hline
		\cite{tang2016modeling} & * & 698 (179 trucks, 519 cars)& * & Extract from 4-month data & Modeling drivers' dynamic decision-making behavior & video-based \\ \hline
		\cite{saitodriver} & 20 & 1 & * & $ \approx $ 4,200 min & Lane departure & DS\\ \hline
		\cite{wang2016driving} & 24 (20 male, 4 female) & 2 & * & 300 min & Car following and cut-in behavior & Field test\\
		\hline
		\hline
	\end{tabular}
	$^ \dagger $All the data listed in this table are from the published papers, where $ \ast $ means that we did not find the accurate information in the references. The driving time $ t $ is the length of experiment time.
	\end{threeparttable}
	\label{list_amount}
\end{table*}

\section{Analysis of Data Size Used in Existing Studies}
As shown in Table \ref{Major_project}, the number of driver participants and the duration used to collect data vary significantly. The differing data amount appearing in the published papers depends greatly on the financial/equipment capabilities of the experiments, the topics focused on, and the the methods employed.
Fig. \ref{TrAc_MDB} and Table \ref{list_amount} show the differences in experimental time\footnote[1]{Experiment time is the duration for conducting an experiment, which differs from the lasting time of driving events. Data collected from the entire period of experiments is called raw data; the data extracted from the raw data is called purified data. The purified data is usually used to model or analyze driver behaviors.} of data collection for research on between traffic accident analysis and modeling driver behaviors. The ``Total time'' includes the time of collecting the raw data or purified data. We do not separate them out, as some references did not clearly distinguish them. The data in Fig. \ref{TrAc_MDB} is collected from 26 published papers. We note that research related to traffic accident analysis generally requires a longer period of time for data collection (about 3 years on average) than research on modeling driver behaviors (about 288 minutes on average). The factors that influence the required amount of NDD for traffic accident analysis are analyzed and discussed. We mainly focus on the required amount of NDD for modeling common driver behavior.

\subsection{Traffic accident analysis}
Traffic accident analysis covers a wide range of topics such as analysis of traffic accident injury severity\cite{de2011analysis,chang2006analysis,delen2006identifying}, relationship analysis between personality and traffic accident\cite{ulleberg2001personality,sumer2003personality,iversen2002personality}, accident hotpots detection or prediction\cite{anderson2009kernel,xie2008kernel}, risk factors analysis\cite{zhang2016traffic}, and traffic accident classification\cite{de2013analysis}. As shown in reference \cite{savolainen2011statistical}, nearly about thirty approaches were applied to traffic accident analysis. Most data in the traffic accident analysis are collected from the local traffic department, recorded and reported by the traffic police, and/or using questionnaire investigation, which usually does not cost so much compared to the naturalistic driving study. But if conducting research on the relationship between the driving styles and traffic accidents based on the NDD, the data collection will cost a great deal. Three main reasons for the traffic accident analysis requiring long running experiments are:

\begin{figure}[t]
	\centering
	\includegraphics[scale = 0.7]{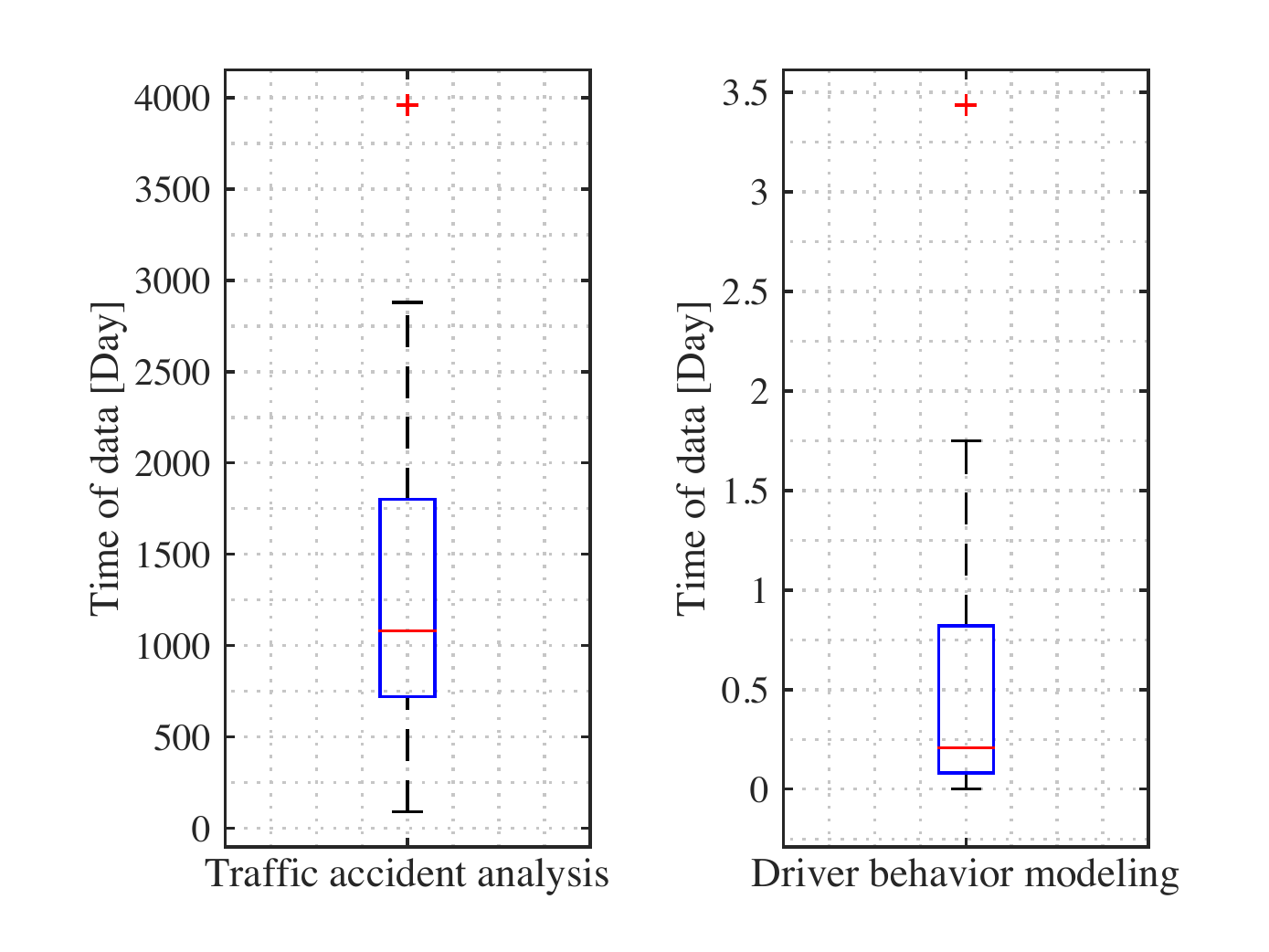}
	\caption{A comparison between the lasting time of data collection for research topics on traffic accident analysis (left) and modeling driver behaviors (right).}
	\label{TrAc_MDB}
\end{figure}

\begin{table*}[t]
	\centering
	\caption{\textsc{Amount of Naturalistic Driving Data for Research on Car-Following (CF) Behavior}$ ^\ddagger $}\label{CF}
	\begin{threeparttable}
		\begin{tabular}{C{0.7cm}|C{0.8cm}|C{0.8cm}|C{1.5cm}|C{5.5cm}|C{5cm}}
			\hline
			\hline
			Ref.&Driver& Event & Time & Methods &Topic\\ 
			\hline
			\cite{Lefevre16} &5 & (600) & 300 min & Gaussian mixture regression \& HMM & Modeling CF behaviors\\ \hline
			\cite{pariota2015linear} & $ \ast $ & 54 & 1173 min & Model-based (Steady-State CF Model) & Modeling CF behaviors \\ \hline
			\cite{koutsopoulos2012latent} & $ \ast $ & 5196 & (45 min) & Latent class model structure & Modeling CF behaviors \\  \hline
			\cite{ossen2006interdriver} & $ \ast $ & 229   & 191 min & Model-based & Interdriver difference\\ \hline
			\cite{higgs2015segmentation} & 20 & 392 & 196 min & Clustering method& Segment driving patterns\\ \hline
			\cite{qi2016appropriate} & 13 & $ \ast $ & 1200 min & Modified latent Dirichlet allocation & Driving style analysis\\ \hline 
			\cite{Khodayari12} & $ \ast $ & 6101 & (45 min) & Neural networks &  Modeling CF behaviors \\ \hline
			\cite{chen2012behavioral} & $ \ast $ & (5000) & 45 min & Model-based (Newell' CF model)& Capturing traffic oscillations\\ \hline
			\cite{miyajima2007driver} & 276 & $ \ast $ & 6 min & GMM and optimal velocity model & Modeling CF behaviors  \\ \hline
			\cite{panwai2007neural} & $ \ast $ & $ \ast $ & 6 min & Neural networks & Modeling CF behaviors \\ \hline
			\cite{jiang2015some} & 25 & 35 & 45 &Proposed a new CF model & Explore features of CF and platoon\\ \hline
			\cite{zaky2015car} & 1 & $ \ast $ & 4.2--5 min & Model-based (Intelligent driver model) & Regime Classification and Calibration\\ \hline
			\cite{jin2014reducing} & $ \ast $ & 5687 & 45 min & Optimization method & Calibrating CF models\\ \hline 
			\cite{ossen2005car} & $ \ast $ & $ \ast $ & 6 min & Model-based (Gazis-Herman-Rothery model) & CF behaviors of individual drivers\\
			\hline
			\hline
		\end{tabular}
		$ ^\ddagger $ All the data is collected from published papers. A value with a bracket indicates that we did not find an accurate value, but we estimated the value using the SPMD datasets. An asterisk $ \ast $ means the reference did not provide any information that can be used to infer the missing value.
	\end{threeparttable}
\end{table*}

\subsubsection{Heterogeneity} The heterogeneity of traffic accidents is reflected in its discretized property in temporal spatial differences. Traffic accident data is generally represented by discrete categories from a variety perspectives. For example, from the  viewpoint of injury severity, traffic accident data can be grouped into different levels such as fatal injury or killed, incapacitating injury, non-incapacitating, possible injury, and property damage only\cite{savolainen2011statistical}. In addition, some heterogeneities of traffic accidents are unobserved, which means that model parameters may vary across observations of traffic accidents. For example, injury severity is likely to exist among the population of crash-involved road users \cite{savolainen2011statistical} because of differences such as risk-taking behaviors or physiological factors. Therefore, to improve the model accuracy and predict the potential a traffic accident,  a huge amount of traffic accident data is normally required.

\subsubsection{Scarcity} Even though the total number of road traffic crashes is high, the rate of these traffic crashes is low in comparison with the number of miles that people drive. Americans drive nearly 3 trillion miles per year \cite{kalra2016driving}, but a failure rate of only 77 per 100 million miles was reported for injuries in 2013. In addition, the diversity in traffic accidents and/or crashes makes a lower rate for a specific kind of traffic accident. For example, the frequency of rear-end crash at the signalized intersection and traffic rush hour will be totally different with the case on the highway. And, different road features and driver's personalities will also cause the diversity in traffic accidents. Therefore, the total number of traffic accidents is high per ten thousands of miles, but for a special or defined case of traffic accident, it is too less to analyze and model this kind of traffic accidents. Thus, to analyze traffic accidents and improve model accuracy, the duration of traffic data should be long enough (usually about 3 years as shown in Fig. \ref{TrAc_MDB}) and cover more kinds of traffic accident events. 
%
%
%
%

\begin{figure}[t]
	\centering
	\includegraphics[width = 0.48\textwidth ]{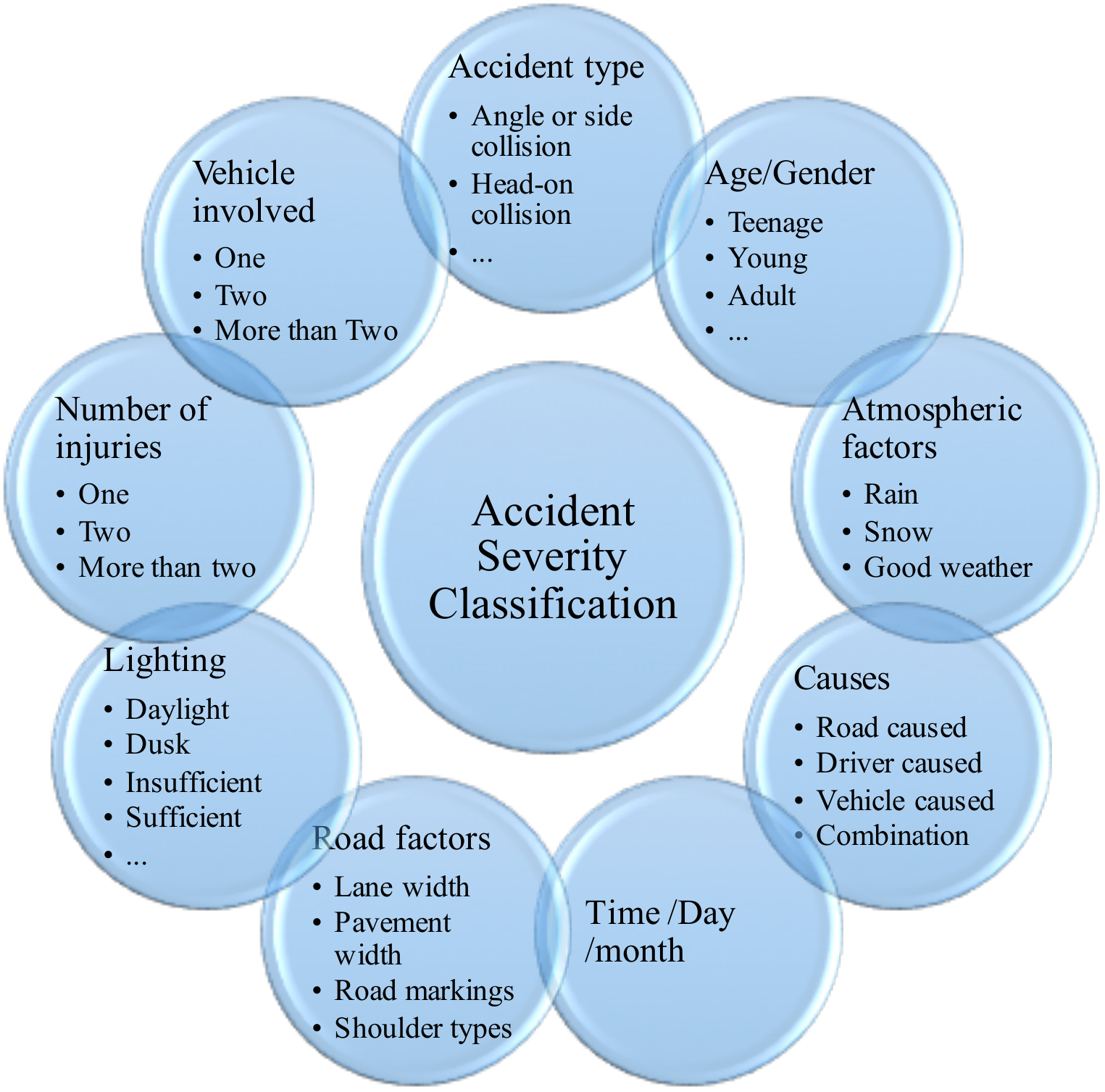}
	\caption{Examples of classifying accident severity based on a variety of criteria.}
	\label{Sec2_classification}
\end{figure}

\subsubsection{Diversity}Traffic accidents can be classified based on criteria such as accident type, age, atmospheric factors, and causes, etc., as shown in Fig. \ref{Sec2_classification}, and also depend greatly on these criteria. Thus,  a more accurate and comprehensive analysis should be based on a great deal of data that would be able to cover nearly all traffic cases yet be sufficient for accounting for all cases of traffic accidents.

Generally, the heterogeneity, scarcity, and diversity of traffic accidents require that the data collection used for traffic accident analysis should cover a long period of time. The time span for collecting data for traffic accident analysis is much longer than that used for understanding and modeling driver behaviors. On the other hand, the cost of data collection for traffic accident analysis is usually lower than the cost related to understanding and modeling driver behaviors because of the different ways of obtaining data. Therefore, in the following section, we discuss and analyze the causes of diversity in the amount of data for modeling driver behaviors.

\subsection{Modeling Driver Behaviors}

Modeling driver behaviors covers a wide range of topics, including, for instance, car following, lane change, left/right turn, U-turn, distraction/inattention, secondary tasks, or brake behaviors. From Fig. \ref{TrAc_MDB}, we know that data for modeling driver behaviors ranges widely from under 50 minutes (e.g, references \cite{chen2012behavioral,miyajima2007driver,panwai2007neural})  to more than 5,000 minutes (e.g., reference \cite{Butakov15}). We present and analyze the reasons for these big differences in terms of \textit{research topic}, \textit{problem formulation method}, and \textit{ data collection methods}. To facilitate the discussion and analysis, we use the car-following behaviors as an example, because car-following behavior is the most common event in driver behaviors.

\subsubsection{Different Research Topics} Table \ref{CF} shows the wide variation in the amount of NDD across research topics on car-following behaviors. For example, some work focused on the microscopic car-following behavior or traffic flow analysis and collected thousands of car-following events\cite{Khodayari12,koutsopoulos2012latent}, while some others focused on individual car-following behavior and applied hundreds of car-following events to research\cite{Lefevre16,higgs2015segmentation}. Moreover, a special case of car-following behavior, i.e., platoon car-following, required more vehicles in the experiment and a higher dimension of driving data for analysis. 

We also found that even for a single kind of research topic, the amount of NDD still varies greatly. For instance, the researchers in\cite{Khodayari12} and \cite{panwai2007neural} used the same method (i.e., neural networks) to model drivers' car-following behaviors, but varied greatly in the amount of data used.

\subsubsection{Problem Formulation Methods} The approach to formulating problems can result in diversity in the amount of NDD. Modeling and analyzing drivers' car-following behaviors, generally involves either a \textit{physically-based} or a \textit{learning-based} method. 
	
\textit{\textbf{(a) Physically-based methods}}: Physically-based method usually describes driver behavior in the form of equations with physical meanings, in which parameters are used to fit the individual driver's characteristics via parameter estimation or calibration methods\cite{zaky2015car,jin2014reducing}. For example, the Gazis-Herman-Rothery (GHR) model describes a driver's car-following behavior by taking current vehicle speed, relative vehicle speed between two adjacent vehicles in the same lane, acceleration, driver reaction time into consideration (see \ref{eq:GHR}).

\begin{equation}\label{eq:GHR}
a_{n}(t) = c\cdot v_{n}^{r}(t)\frac{\Delta v(t-T)}{\Delta x^{l}(t-T)}
\end{equation}
where $ a_n $ is the acceleration of vehicle $ n $; $ v_n^r $ is the speed of the $ n $th vehicle, $ \Delta x $ and $ \Delta v $ are the relative spacing and speeds, respectively, between the $ n $th and $ n - 1 $ vehicle (the vehicle immediately in front) at an earlier time $ t - T $; $ T $ is the driver reaction time; $ r $, $ l $ and $ c $ are the constants to be determined. Most popular car-following models, including the GHR model, intelligent driver model, optimal velocity model, and collision avoidance models, were compared and evaluated in \cite{brackstone1999car,panwai2005comparative}. Thus, the requisite amount of data depends on a number of unknown parameters in physical models. Generally speaking, a physical model with many unknown parameters requires more driving data to fit driver behaviors. In addition, the amount of required data also depends on the method used to calibrate car-following models. For example, a calibration method using statistical techniques usually requires more data than that without considering the statistical features.

\textit{\textbf{(b) Learning-based methods}}: Learning-based methods utilize machine learning techniques, without considering the physical meaning of the model parameters, to describe more complex and underlying nonlinear relationships between different kinds of surrounding traffic information and driver behaviors. Due to the complexity and diversity of drivers' car-following behaviors, it is generally difficult to capture the stochastic features of drivers using physically-based model. A learning-based method is therefore introduced to solve these kinds of issues. For example, neural networks \cite{Khodayari12,panwai2007neural}, a Gaussian mixture regression--hidden Markov model \cite{Lefevre16,wang2017learning} and recurrent neural networks \cite{morton2016analysis} have been applied to modeling, analyzing and characterizing driver behaviors. 
Therefore, different types of problem formulation require different amount of data.

	\begin{figure}[t]
		\centering
		\begin{subfigure}{0.48\textwidth}
			\centering
			\includegraphics[width = 1\textwidth]{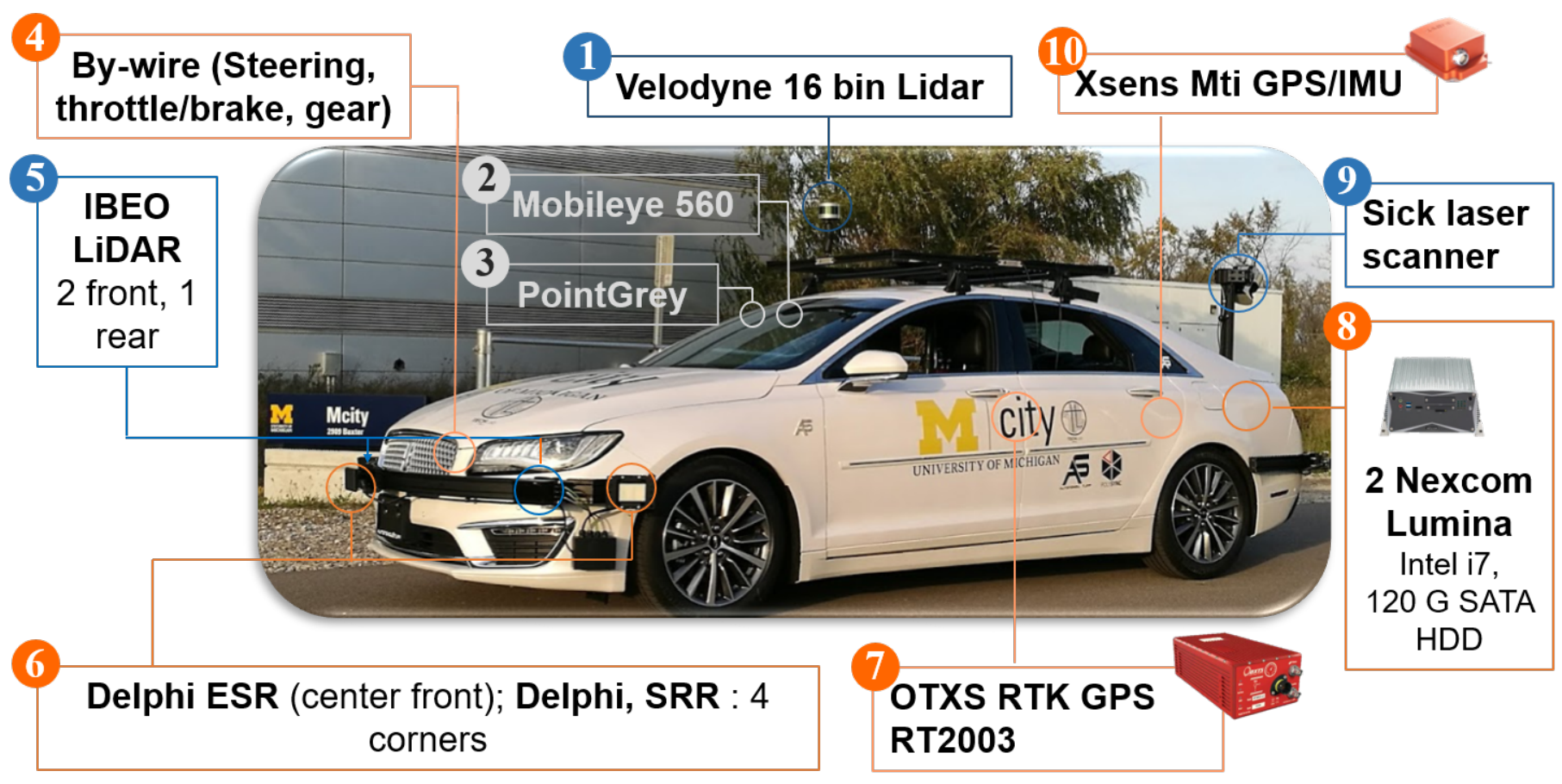}
			\caption{Example of in-vehicle data acquisition systems developed by University of Michigan.}
		\end{subfigure}
		\begin{subfigure}{0.48\textwidth}
			\centering
			\includegraphics[width = 1\textwidth]{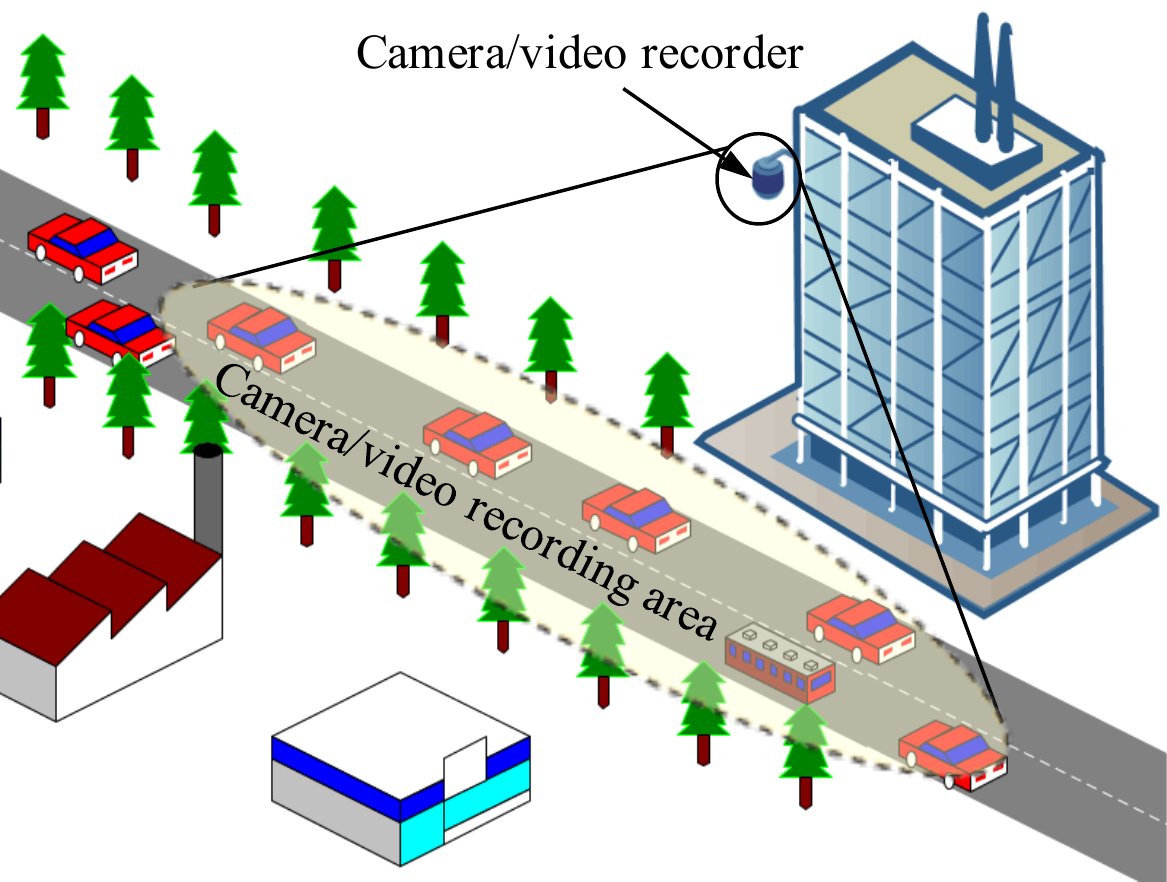}
			\caption{Illustration of data acquisition systems for car-following behaviors using a camera/video recorder with a fixed position.}
		\end{subfigure}
		\caption{Illustrations of two different data collection methods.}
		\label{fig:DAS}
	\end{figure}

\subsubsection{Data Collection Approaches} The approach to collecting driving data  varies across research topics. Past data collection approaches included: \textit{in-vehicle sensor data} and \textit{video/camera data with a fixed field} (Fig. \ref{fig:DAS}).

	\begin{figure}[t]
		\centering
		\includegraphics[scale = 0.8]{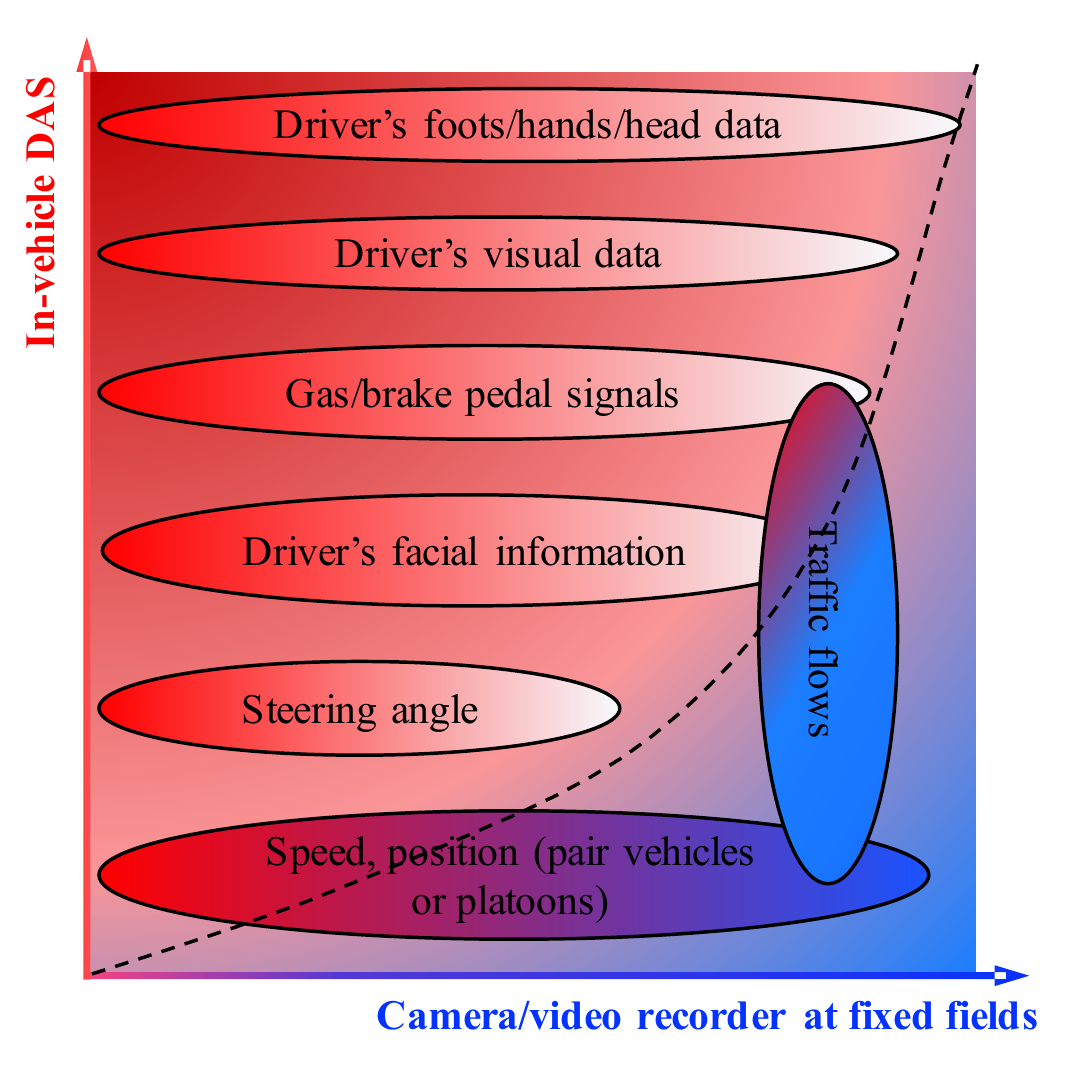}
		\caption{The illustration of information that could be collected from two different methods.}
		\label{fig:DAS_inf}
	\end{figure}
	
	\textit{\textbf{(a) In-vehicle sensor data}}: The NDD collected from in-vehicle sensors, such as cameras and/or radar that can sense information about adjacent vehicles in the same lane and driver's personality, is referred as in-vehicle sensor data. Fig. \ref{fig:DAS}(a) shows an example of an in-vehicle data acquisition system developed by the University of Michigan which consists of an array of sensors such as laser scanners, cameras, and Lidars. For example, Wang \cite{wang2013adaptive} \textit{et al}. used cameras to monitor the road, the driver's foot as well as steering hands and analyzed a driver's car-following characteristics. Higgs and Abbas \cite{higgs2015segmentation} collected the NDD based on in-vehicle cameras, radars, and CAN-Bus signals to analyze a driver's car-following patterns. In addition, the high-precision difference in GPS devices (e.g., Multi-functional Satellite Augmentation System, a product from Japan) can also be directly used to record vehicle speed and position, which can be applied to a pair of cars or car-platoon behaviors \cite{jiang2015some}. Currently, most data acquisition systems on the market, such as Mobileye used in SPMD program\cite{bezzina2014safety} and the data acquisition system in SHRP 2 program developed by VTTI \cite{campbell2012shrp} , can be reliably used to collect driving data. This kind of in-vehicle equipment or data acquisition system costs are high, and thus most researchers can not afford a complete set of data acquisition system. Data obtained via the in-vehicle data acquisition system may include data of driver actions/behaviors (e.g., eyes detection, hands detection, and foot action), road features (e.g., road curvature, road/lane width), information of front vehicles (e.g., relative distance, relative speed) and ego vehicle data through CAN-Bus (e.g., acceleration, vehicle speed, throttle opening, steering angle). Thus, for an individual driver, a vehicle with this kind of data acquisition system can be used to built driver behavior models, analyze driver distraction/inattention, ascertain the decision-making process and personal characteristics, and drivers' visual-cognitive, physical and psychomotor capabilities. If many drivers were involved, studies on the difference across individuals could also be conducted, but at a much higher cost.
%
%
%
%
	
	\textit{\textbf{(b) Video/camera data with a fixed field}}: A lower cost alternative but efficient way is to install a video recorder at a fixed position, obtaining video-based data  (e.g., vehicle trajectories and positions) to analyze driver behaviors, as shown in Fig. \ref{fig:DAS}(b). This approach has been widely used to collect vehicle trajectory data and analyze traffic flows or build the car-following model. For example, Yu \cite{yu2014extended} \textit{et al}. collected the car-following data by installing a video recorder on the windowsill of a tall building adjacent to the intersection, and then utilized these data to analyze the influencing factors of car-following behaviors at urban signalized intersections, determining the structure of an extended car-following model. Some researchers also fixed the camera/video recorder on a helicopter\cite{ozaki1993reaction,ossen2006interdriver}, traffic light signal poles and structures to collect driving data. This kind of data collection method allows researchers to obtain a huge amount of driving data for many vehicles at a lower cost and with less time, though tracking a single driver's other behaviors, such as steering angle, head movement, and eye information, is difficult. For instance, more than 6 thousand vehicle trajectories in \cite{jin2014reducing} take the researchers only about 45 minutes to obtain using this method, but included no data on steering angle, head movement. While the method based on an in-vehicle data acquisition system records high-dimension data (Fig. \ref{fig:DAS_inf}), it is very difficult to obtain so many vehicle trajectories of car-following events in a short period of time. As such, this method is usually used for developing a car-following model and analyzing car-following behaviors from a general viewpoint.

Fig. \ref{fig:DAS_inf} summarizes and presents the comparisons between two approaches of data collection. We note that the collection approach using in-vehicle data acquisition systems, compared to camera/video recorder at a fixed field, can collect a wide range of data from the driver's foot movement to vehicle velocity. The method based on a fixed field camera/video recorder, is  best used for collecting a large amount of driving data (i.e., different vehicles) but covering fewer types data.

A video/camera  in a fixed field can collect a great amount of driving data at a lower cost, but the diversity of data limits its application in deeply understanding and modeling driver behavior. Thus, most researchers would prefer to utilize multivariate in-vehicle sensors even if it costs more. In the next section, we propose and show a general approach to determine the appropriate amount of NDD for modeling driver behaviors based on an in-vehicle data acquisition system.

\section{Proposed Methods}
We present an analysis tool to determining how much NDD collected from in-vehicle sensors is sufficient from a statistical point of view. Our proposed methods focus mainly on determining how much NDD is enough to cover the features of driver behaviors rather than assessing which method is better for modeling driver behaviors.

\subsection{Why a Statistical Method?}
As discussed in Section II, the amount of NDD varies greatly due to the diversity of research topics, data collection methods, and problem formulation approaches. To develop a flexible approach, we make two assumptions as follows:
%
%
\begin{itemize}
	\item A better driver model or an analysis of driver behavior characteristics should be based on a set of NDD that can cover almost all of the driver's basic characteristics. As such, a driver model built on, or driving characteristics inferred from, an insufficient data set are not suitable for applications.
	\item Driver behavior is highly affected by uncertainty caused by the surroundings (e.g., other road users) and the driver themselves (e.g., their emotions and mental states), but over the long period of time of driving, the statistical characteristics of driving behavior for an individual driver will be convergent\cite{sagberg2015review,hakkinen1958traffic}. Namely, a driver will adapt to himself/herself driving styles and then finally shape a stable driving style according to his/her internal model after a long-time period of driving.
\end{itemize}
In line with the above assumptions, we estimate the appropriate amount of data by finding the convergent point of the density function of collected data from a statistical perspective. The distribution of the NDD sequence $ \bm{x} = \{x_{i}\}_{i=1}^{n} $  is estimated and denoted as $ \hat{F}(x;n) $, and its density is $ \hat{f}(x;n) = \frac{d}{dx} \hat{F}(x;n) $ under $ n $ observations. For different observation amounts $ n $, the density of observations $ \hat{f}(x;n) $ will be different. If an adequate amount of data is provided, the density of observations $ \hat{f}(x;n) $ should change slightly with $ m $ additional observations, i.e., 

\begin{equation}
\hat{f}(x;n) \sim \hat{f}(x;n+m), \ \mathrm{with} \ n \rightarrow \infty, \ m\in \mathbb{R}^{+}
\end{equation}

If adding more observations does not change the distribution, we consider the additional data is redundant. Thus, we treat the $ n $ amount of data as suitable from the statistical perspective, because: (1) the $ n $ amount of data can cover almost all of the underlying characteristics of driver behaviors and (2) adding more data can not provide more useful information. The estimated method of density $ \hat{f}(x;n) $ is presented formally below.

\subsection{Univariate Kernel Density Estimation}
Driver behavior data can be formulated using a parametric method such as a multivariate Gaussian mixture model (GMM) \cite{Butakov15,miyajima2007driver,Lefevre16}. It is difficult, however, to directly assess the similarity of two multivariate GMMs, particularly when the number of GMM components is big. In this paper, we utilize a non-parametric method, that is, kernel density estimation (KDE) method, to estimate the density for a given data sequence. 

Given a sampling dataset $ \{x_{i}\}_{i = 1}^{n}  $ with density function $ f(x) $, the estimated density from the data sample $ x $ can be formulated by\cite{bishop2007pattern}

\begin{equation}\label{eq:kernel_fun}
\hat{f}(x;n) = \frac{1}{n}\sum_{i=1}^{n}\frac{1}{h^{D}}\cdot\kappa\left( \frac{x-x_{i}}{h} \right) 
\end{equation}
where  $ h $ is the bandwidth, $ \kappa(u) $ is the kernel function and a Gaussian kernel function is selected, i.e., $ \kappa(u)  = 1/\sqrt{2\pi}\cdot\exp (-u^{2}/2)$. Thus, we can generate a density function $ \hat{f}(x;n) $ on the basis of a given data sample $ \bm{x} $ with $ n $ observations. During the kernel density estimation, the kernel bandwidth $ h $ has a great influence on the estimated kernel function. A large kernel bandwidth $ h $ will result in an over-smooth issue and inversely a small kernel bandwidth $ h $ will cause an under-smooth issue. In this paper, we applied a Gaussian kernel function with the bandwidth can be estimated by $ h = 1.06\cdot \hat{\sigma}\cdot n^{-1/5} $ \cite{silverman1986density}, where $  \hat{\sigma} $ is the standard deviation of the training data $ \{x_{i}\}_{i = 1}^{n} $.

\begin{figure}[t]
	\centering
	\includegraphics[scale = 0.7]{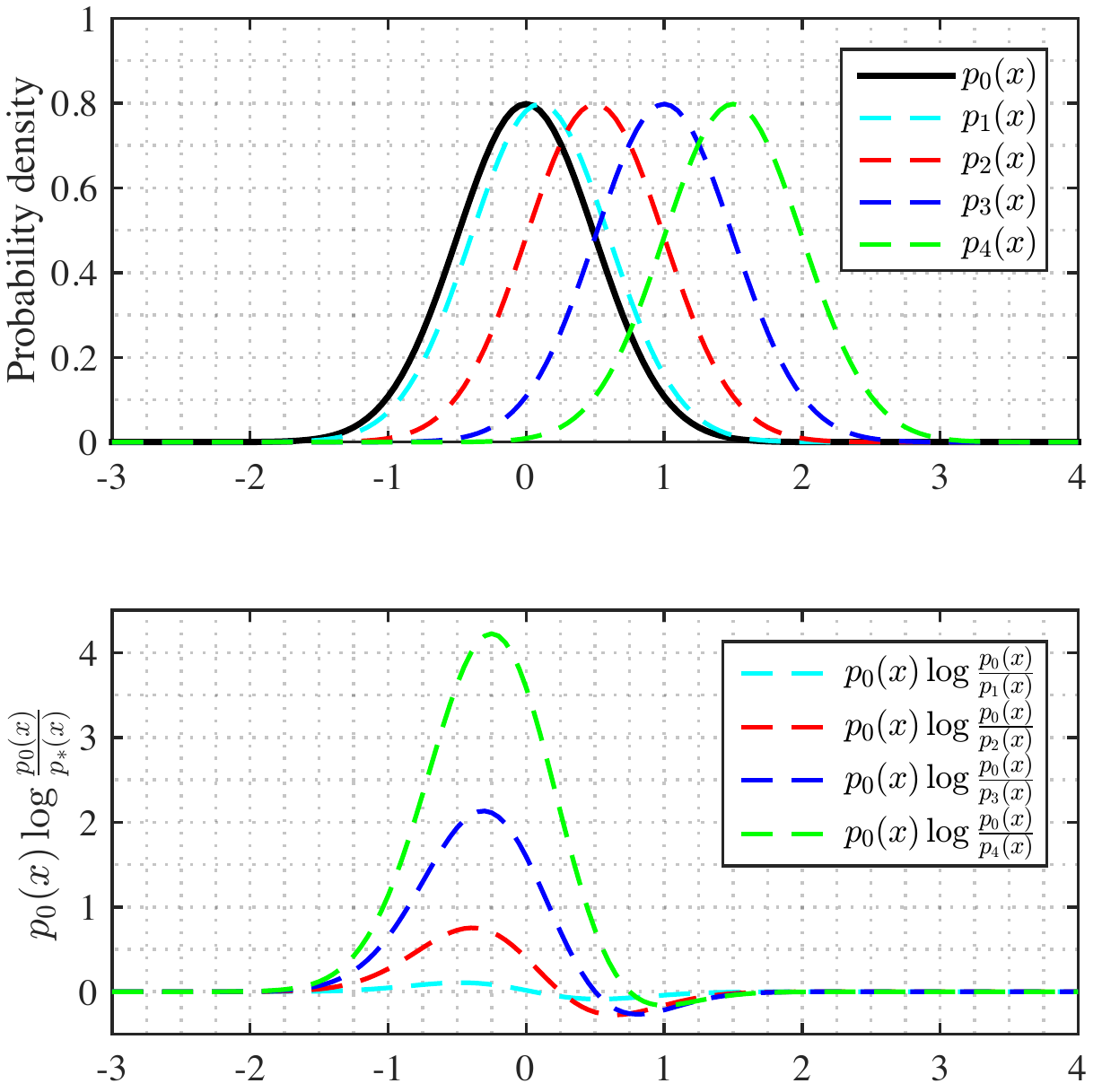}
	\caption{Illustrations of the integral term ($ f\log (\frac{f}{g}) $). Top: different density distributions. Bottom: the values of integral terms for different distributions.}
	\label{KL_example}
\end{figure}

\subsection{Kullback-Liebler Divergence}

We will assess the similarity between two adjacent kernel functions estimated from $ n $ and $ n+m $ data observations. To achieve this, we employ the Kullback-Liebler (KL) divergence index \cite{bishop2007pattern} to test the similarity between the distribution of two adjacent data sets, defined by

\begin{equation}\label{eq:KL}
\begin{split}
KL\left( \hat{f}(x;n+m)||\hat{f}(x;n)\right)  = \int & \bigg[  \hat{f}(x;n+m) \\ & \times \log\frac{\hat{f}(x;n+m)}{\hat{f}(x;n)}\bigg]
\end{split}
\end{equation}
The $ KL $ can quantify the level of similarity between two density functions as follows:
\begin{enumerate}
	\item when $ KL(\hat{f}(x;n+m)||\hat{f}(x;n)) $ approaches 0, it indicates that $ \hat{f}(x;n) $ is extremely close to $ \hat{f}(x;n+m) $, meaning that additional data would not provide more useful information to the density function;
	\item when $ KL(\hat{f}(x;n+m)||\hat{f}(x;n))$ becomes large, it indicates that $ \hat{f}(x;n) $ is different from $ \hat{f}(x;n+m) $, indicating that more data is needed.
\end{enumerate}
Fig. \ref{KL_example} provides an example to illustrate the  KL  divergence between different normal density functions. The top picture shows five normal density distributions with different center values, where the black line represents the basic density function. The bottom picture shows the values of the integral term in (\ref{eq:KL}) between the other four density functions and the basic density function. We note that (1) when the probability density $ p_{0}(x) $ is close to $ p_{1}(x) $, the sum value of $ p_{0}(x)\log (\frac{p_{0}(x)}{p_{1}}) $ approaching to zero and (2) when the probability density $ p_{0}(x) $ is different  from $ p_{4}(x) $, the sum value of $ p_{0}(x)\log (\frac{p_{0}(x)}{p_{4}}) $ becomes larger.

We thus determine the proper amount of driving data so that $ KL(\hat{f}(x;n+m)||\hat{f}(x;n)) $ change very slightly, even if more data samples were to be added, i.e.,

\begin{equation}\label{eq:KL_critical}
\begin{split}
& \bigr| KL(\hat{f}(x;n+m)||\hat{f}(x;n)) - \\ & KL(\hat{f}(x;n+2m)||\hat{f}(x;n+m))\bigl|  \leq \epsilon,  \  \epsilon \in \mathbb{R}^{+}  \\
\end{split}
\end{equation}
where $ \epsilon $ is a small positive value. It is obvious that a larger value of $ \epsilon $ can lead to a small amount of the required NDD. In this paper, to obtain a more conservative result, we set $ \epsilon = 10^{-4} $. 

\section{Case Study of Modeling \\Driver Behaviors}
The NDD has been widely used to extract, model, and understand driver behaviors or their internal mechanisms, as a new way to design vehicles that transition from automated to manual driving \cite{russell2016motor}, to develop personalized driver assistance systems \cite{Butakov15,wanghuman,Butakov16,wang2017learning}, and to improve fuel efficiency\cite{ferreira2015impact} as well as vehicle/road/traffic safety\cite{sagberg2015review}. However, the stochastic features and nonlinearity of driver behaviors make it difficult to directly model and analyze driver behaviors as dynamical systems \cite{Butakov15}. A more efficient way is to treat driver behaviors as a stochastic process and fit a model or extract features from a large quantity of data, called the data-driven method. Driving data can be collected using four different testing approaches\cite{karl2013driving}: (1) driving simulators, (2) quasi-experimental field studies, (3) field operational tests and (4) naturalistic driving studies. Compared to the first three methods, driving data collected from the fourth method (i.e., NDD) can more accurately reflect a driver's natural traits, but they are very costly and time intensive \cite{karl2013driving,akamatsu2013automotive}. An appropriate amount of NDD is required to avoid insufficient or excessive data to save time and money and to improve model accuracy. In this section, we investigate and answer the question ``\textit{How much naturalistic driving data is enough to model drivers' behaviors?}'' by taking the case of modeling car-following behaviors as an example. 

\begin{figure}[t]
	\centering
	\includegraphics[width = 0.48\textwidth]{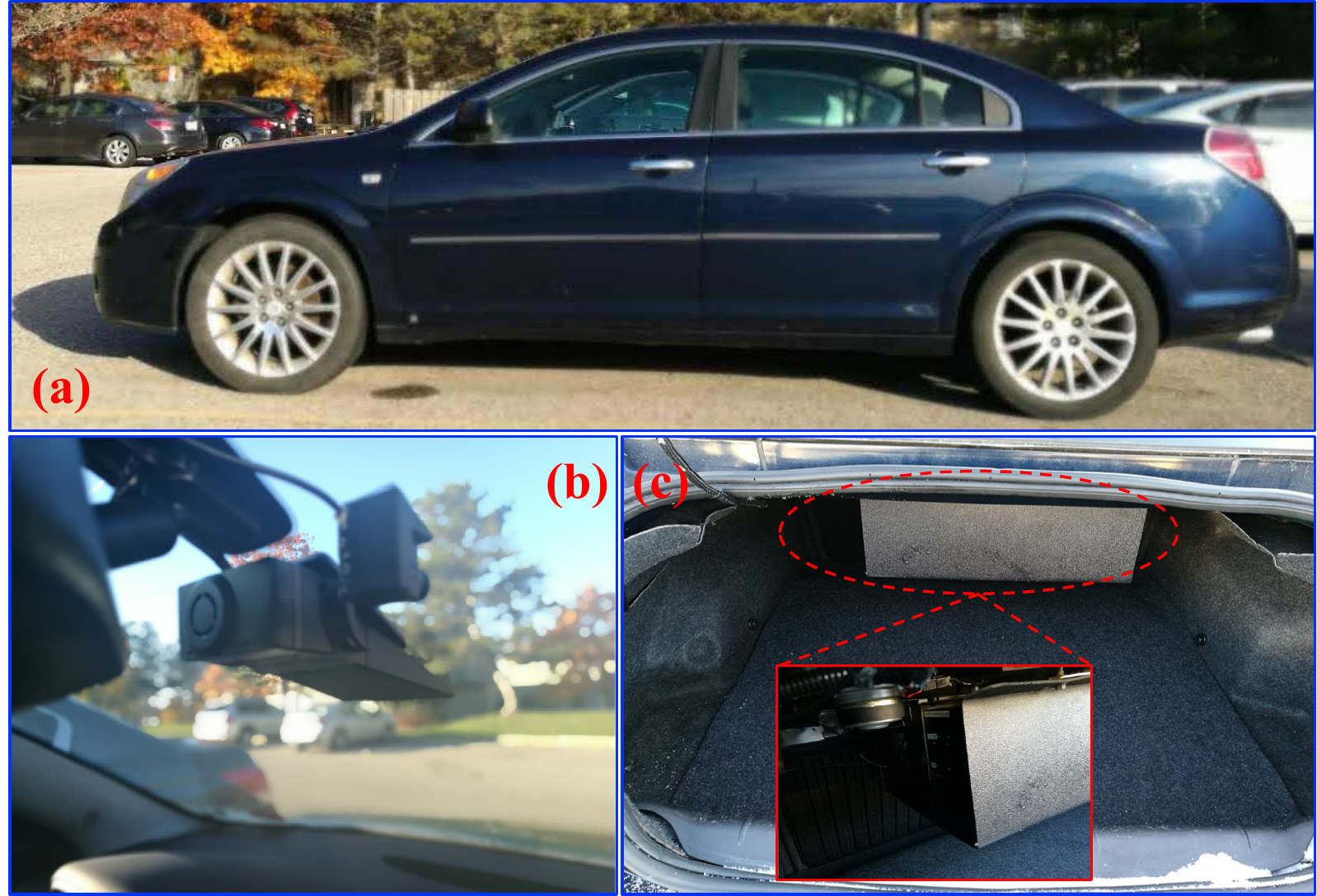}
	\caption{An example of the data collection equipment: (a) Experiment vehicle; (b) Mobileye; (c) Data acquisition system.}
	\label{fig:equipment}
\end{figure}

\subsection{Experiments}

\begin{figure}[t]
	\centering
	\includegraphics[width = 0.48\textwidth]{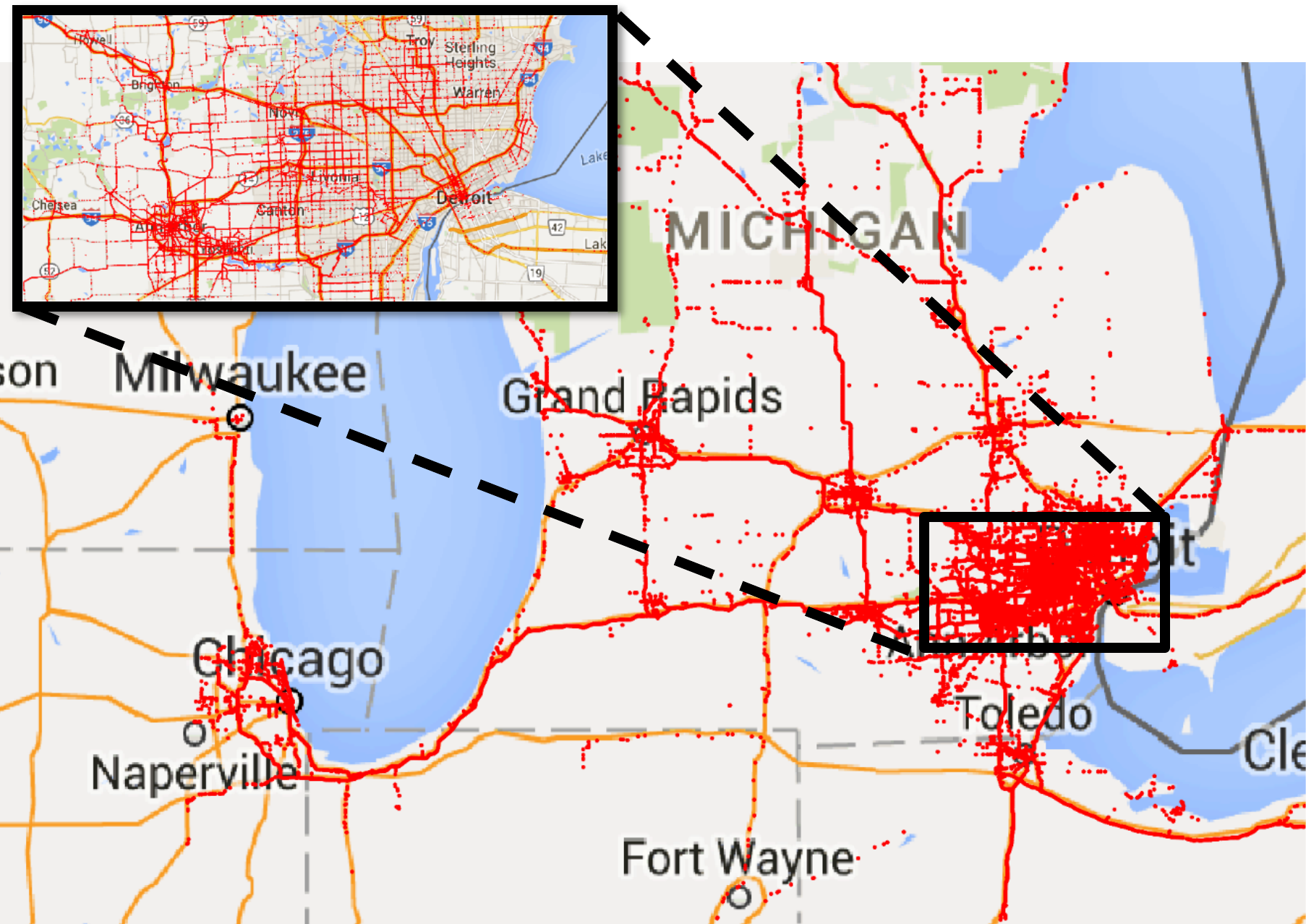}
	\caption{The trajectories of all car-following data.}
	\label{fig:datatrips}
\end{figure}
The NDD used in this research was extracted from the SPMD database. It recorded the naturalistic driving of 2,842 equipped vehicles in Ann Arbor, Michigan, for more than two years. As of April 2016, 34.9 million miles were logged, making the SPMD one of the largest public naturalistic fields of test databases ever. We used 98 sedans to run experiments and collect the real on-road data. The experiment vehicles were equipped with a data acquisition system and MobilEye, as shown in Fig. \ref{fig:equipment}. The in-vehicle data includes vehicle speed, acceleration, and GPS signal from the CAN-bus. The lateral position with respect to lane or road edges were recorded by MobilEye. All driver participants had an opportunity to drive in rural, urban, and highways situations without any specific restrictions or requirements, as shown in Fig. \ref{fig:datatrips}. The NDD were recorded at the rate of 10 Hz or 10 samples per second. 

\subsection{Driving Scenarios Definition}

\begin{figure}[t]
	\centering
	\includegraphics[width = 0.48\textwidth]{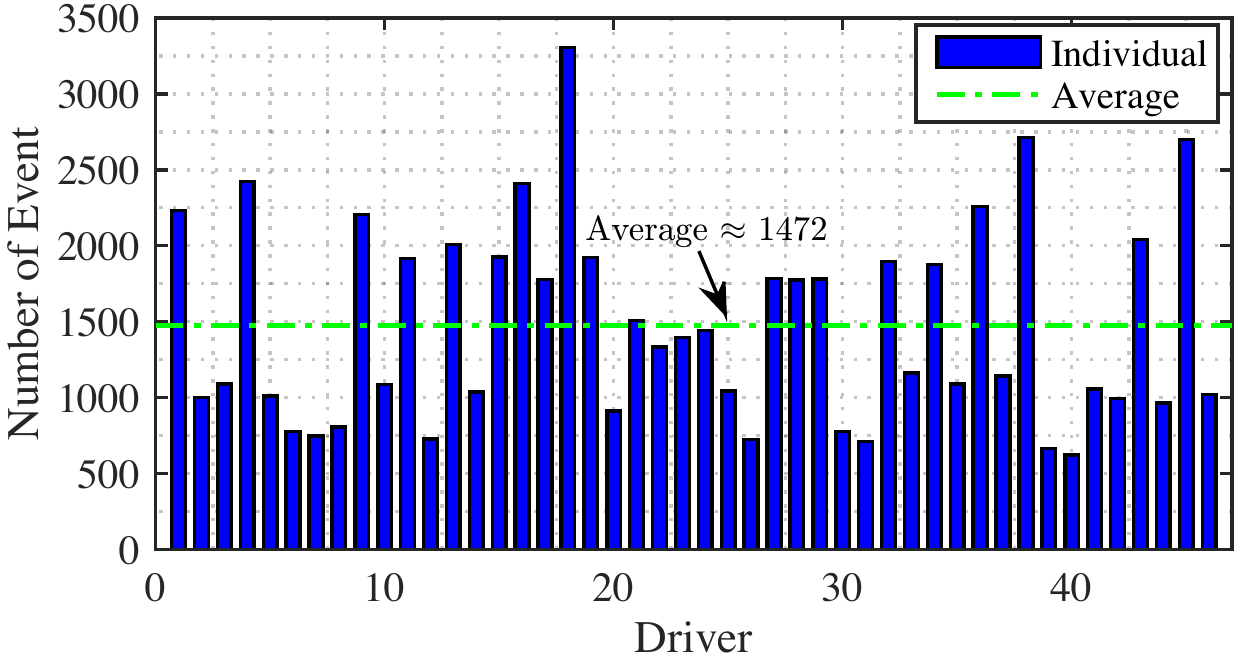}
	\caption{Statistical information of NDD for 46 drivers.}
	\label{fig:Data_Inf}
\end{figure}

We define the following variables to describe drivers' car-following behavior between two adjacent vehicles in the same lane. The ego vehicle is the vehicle we model. The preceding vehicle is the adjacent vehicle located ahead in the same lane as the ego vehicle. To extract the data from the entire database, we define the car-following scenario as follows:
\begin{enumerate}
	\item Ego vehicle is close to the preceding vehicle in the same lane. The relative distance between the ego vehicle and the preceding vehicle must be longer than 120 m\cite{higgs2015segmentation}. If the relative distance between the two vehicles is larger than 120 m, this driver behavior was treated as a free-following case.
	
	\item The speed of the ego vehicle is larger than 5 m/s. The limitation is placed on speed to separate the car-following data from the traffic jam data and Stop\&Go data.
	
	\item The cut-in behavior of surrounding vehicles or lane change behavior of the ego vehicle is also not involved. When a car cut-in  from the neighboring lane to the gap between the current preceding vehicle and the ego vehicle, or the ego vehicle makes a lane change behavior, the car-following event will end.
	
	\item The length of the car-following period must be greater than 30 s\cite{higgs2015segmentation}, and the number of car-following events for each driver should be larger than 300. The two limitations ensure that the NDD is sufficiently large for determining the appropriate amount.
\end{enumerate}
After data extraction, most typical car-following behavior were included such as data related to constant moving speed of the leading car at various speed, data related to constant acceleration, deceleration, oscillation with various amplitude and frequency, etc.

\begin{figure}[t]
	\centering
	\includegraphics[width = 0.48\textwidth]{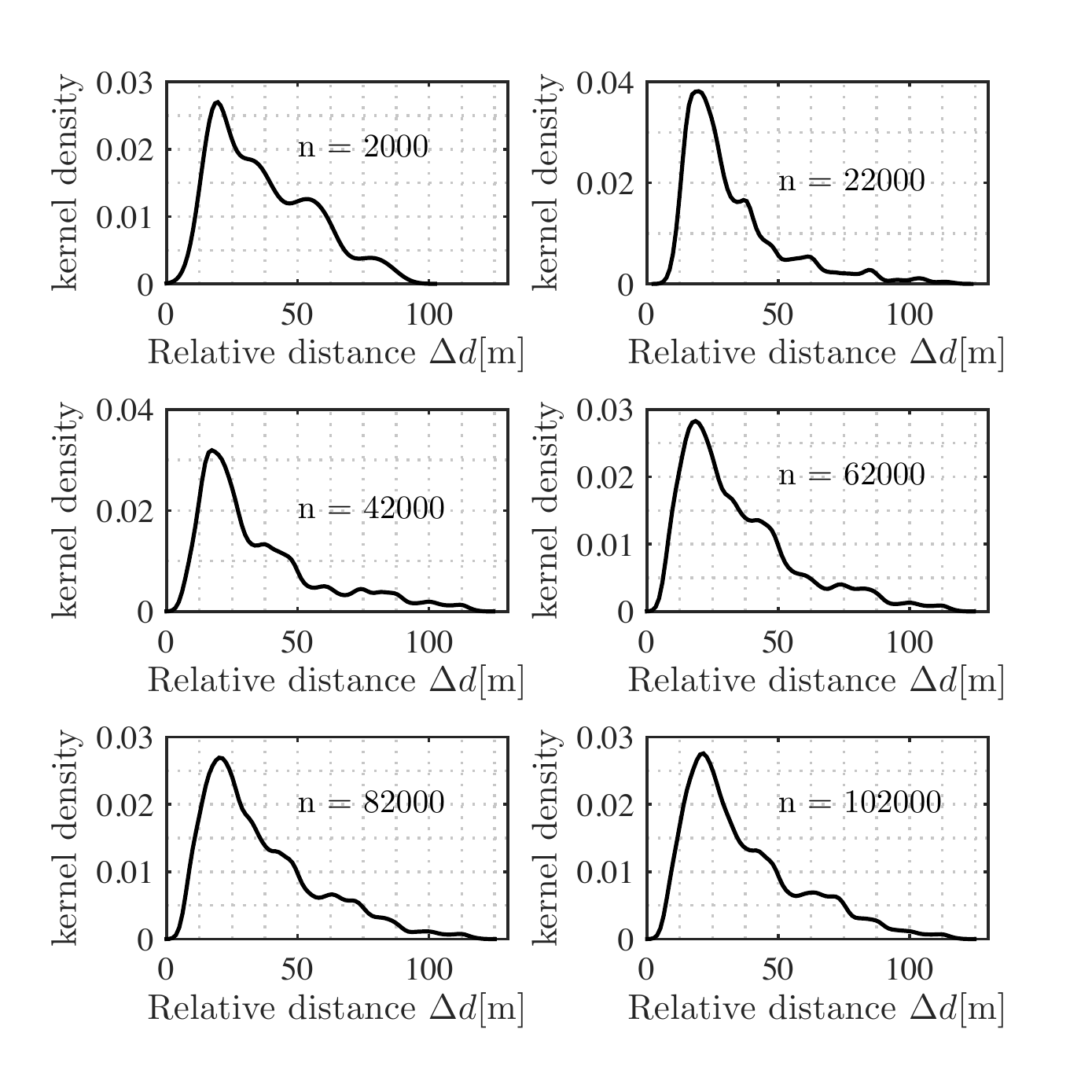}
	\caption{Example of kernel density of relative distance for driver \#12 car-following behavior using different amounts of NDD.}
	\label{fig:kernel_RelDis}
\end{figure}
\begin{figure}[t]
	\centering
	\includegraphics[width = 0.48\textwidth]{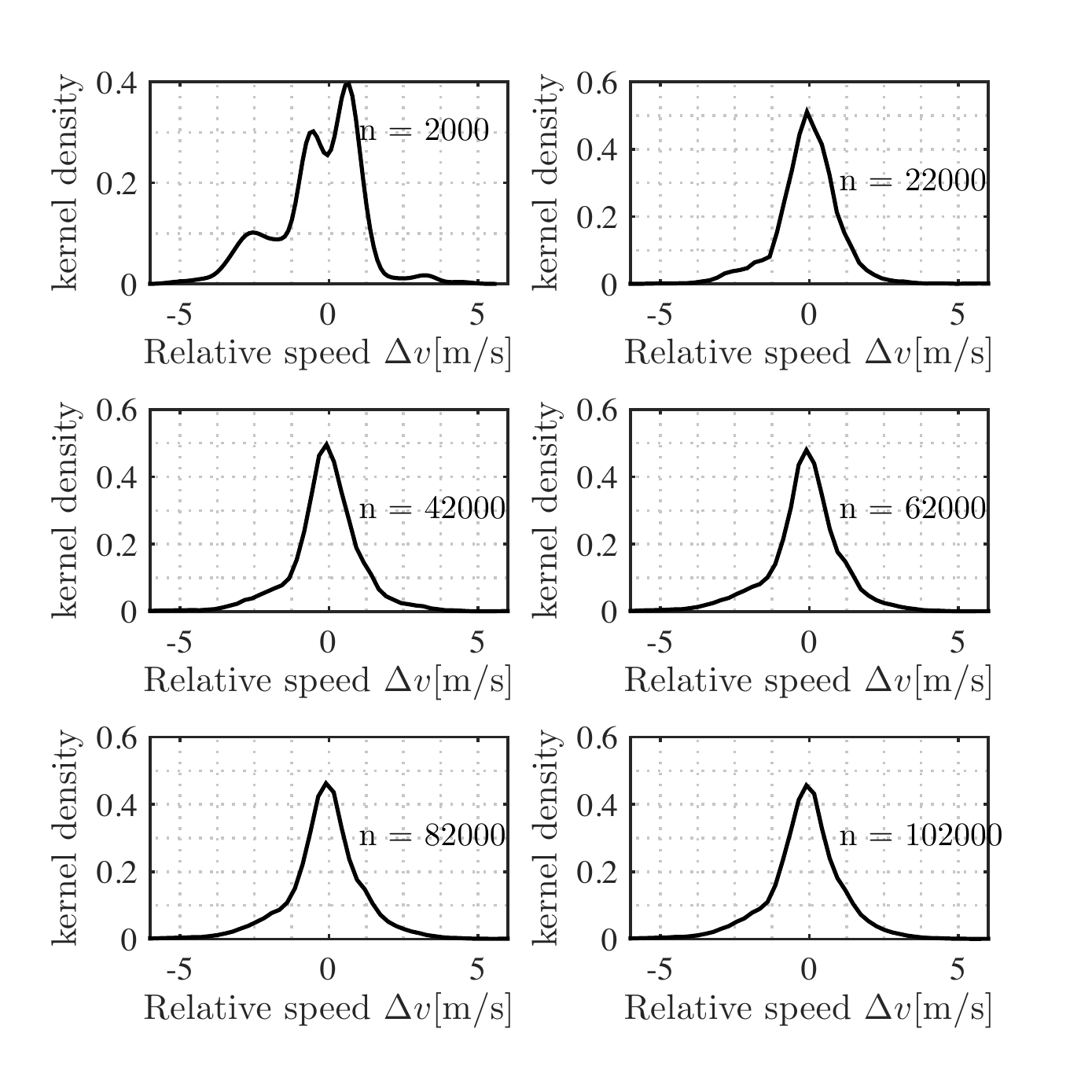}
	\caption{Example of kernel density of relative speed for driver \#12 car-following behavior using different amounts of NDD.}
	\label{fig:kernel_RelSp}
\end{figure}
\begin{figure}[t]
	\centering
	\includegraphics[width = 0.48\textwidth]{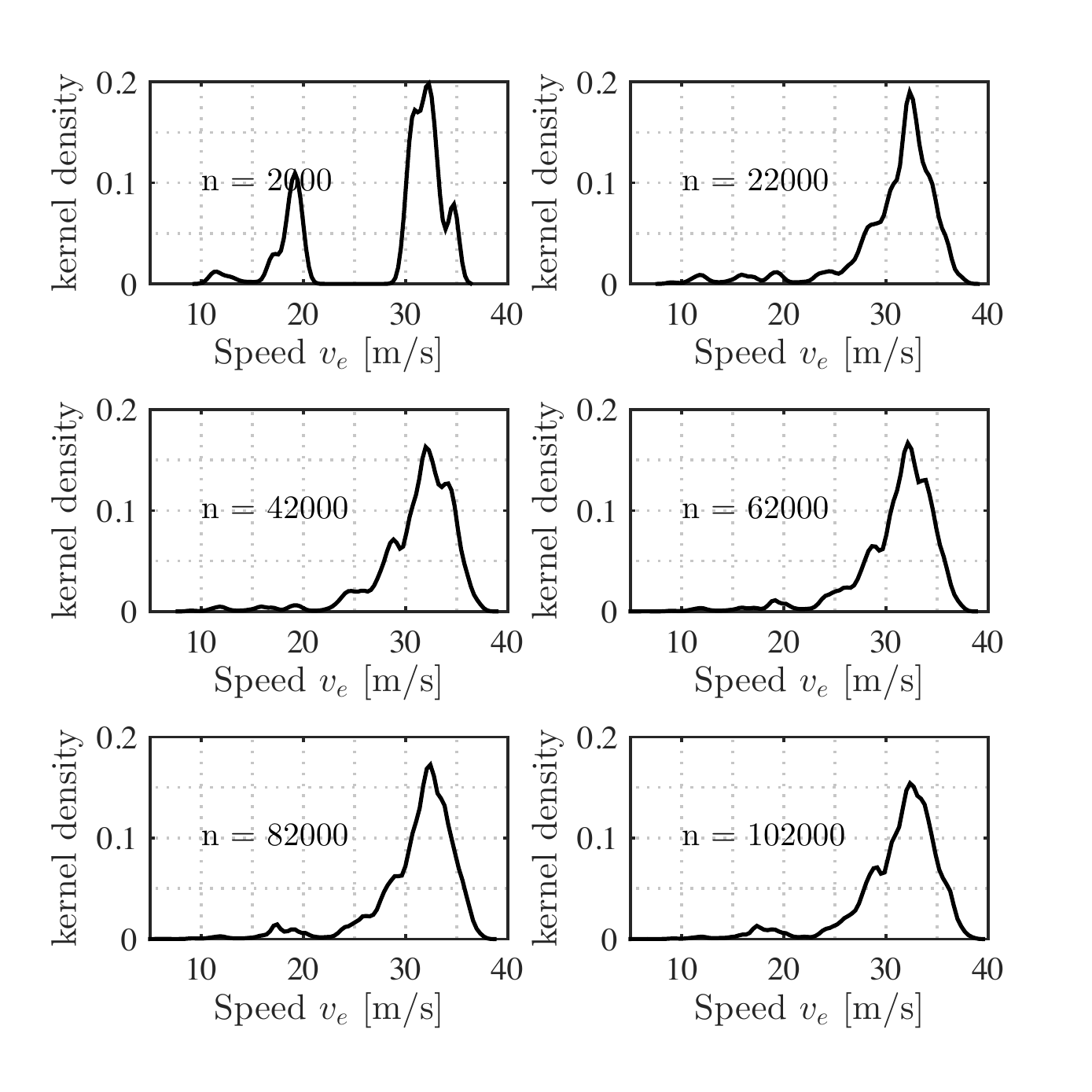}
	\caption{Example of kernel density of speed for driver \#12 car-following behavior using different amounts of NDD.}
	\label{fig:kernel_Sp}
\end{figure}
\begin{figure}[t]
	\centering
	\includegraphics[width = 0.48\textwidth]{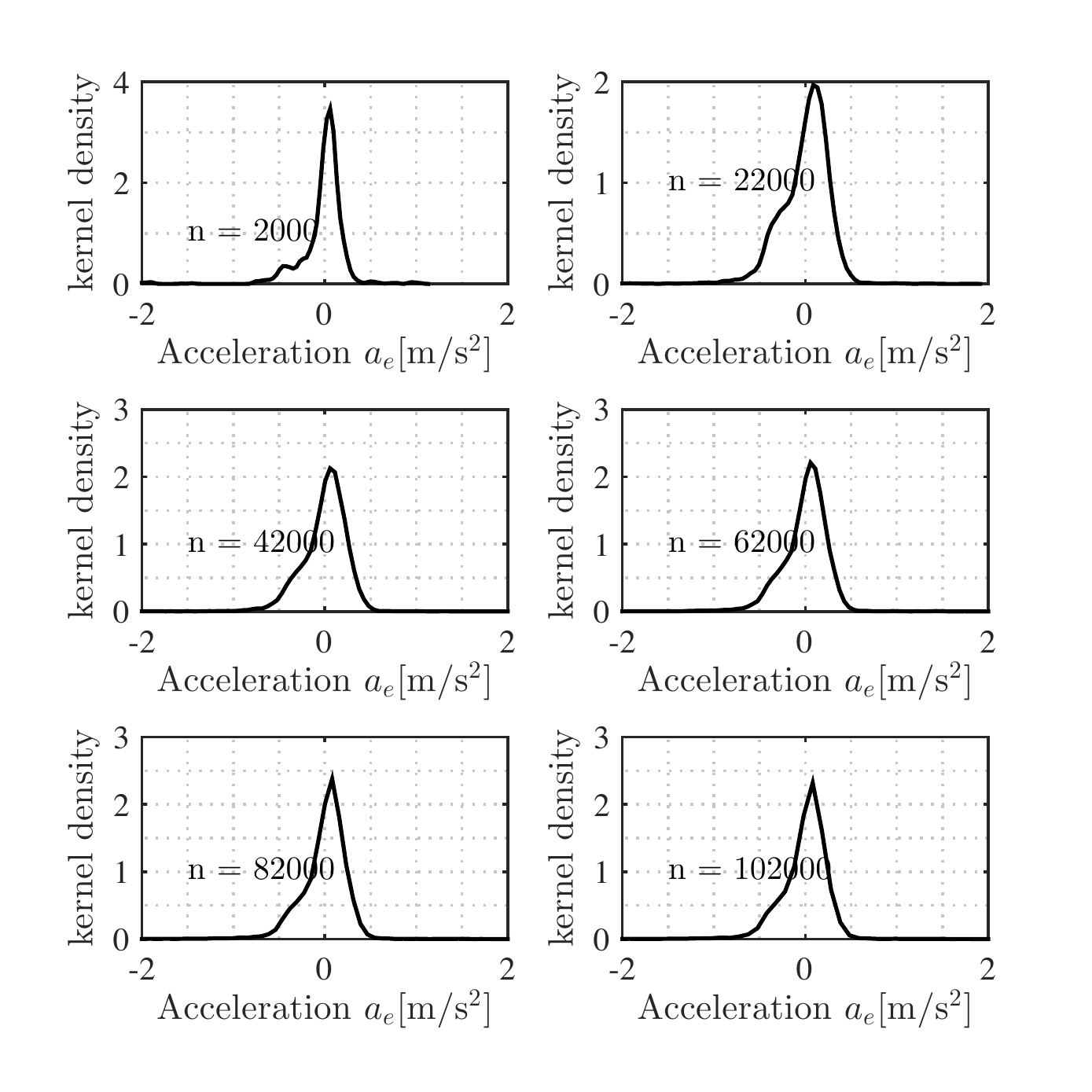}
	\caption{Example of kernel density of acceleration for driver \#12 car-following behavior using different amounts of NDD.}
	\label{fig:kernel_acc}
\end{figure}

\begin{figure*}[t]
	\centering
	\begin{subfigure}[t]{0.48 \textwidth}
		\centering
		\includegraphics[scale = 0.66]{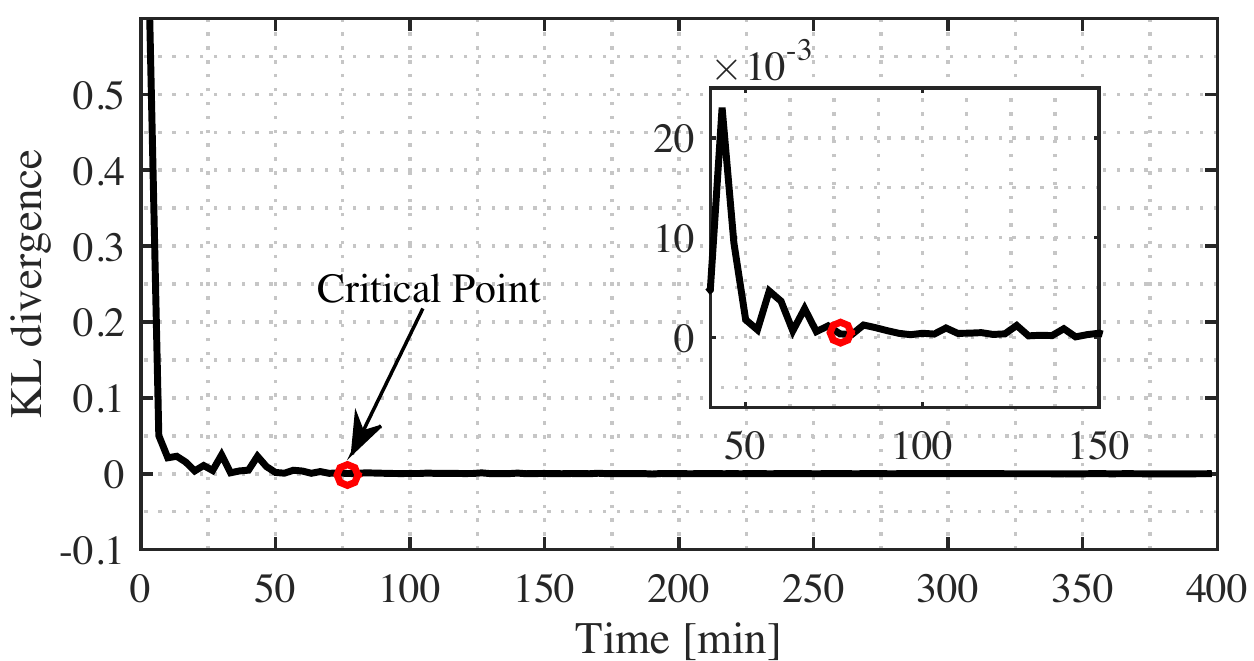}
		\caption{Relative range, $ n^{\ast}_{\Delta d} \approx 4.6 \times 10^{4}$ }
	\end{subfigure}
	~
	\begin{subfigure}[t]{0.48 \textwidth}
		\centering
		\includegraphics[scale = 0.66]{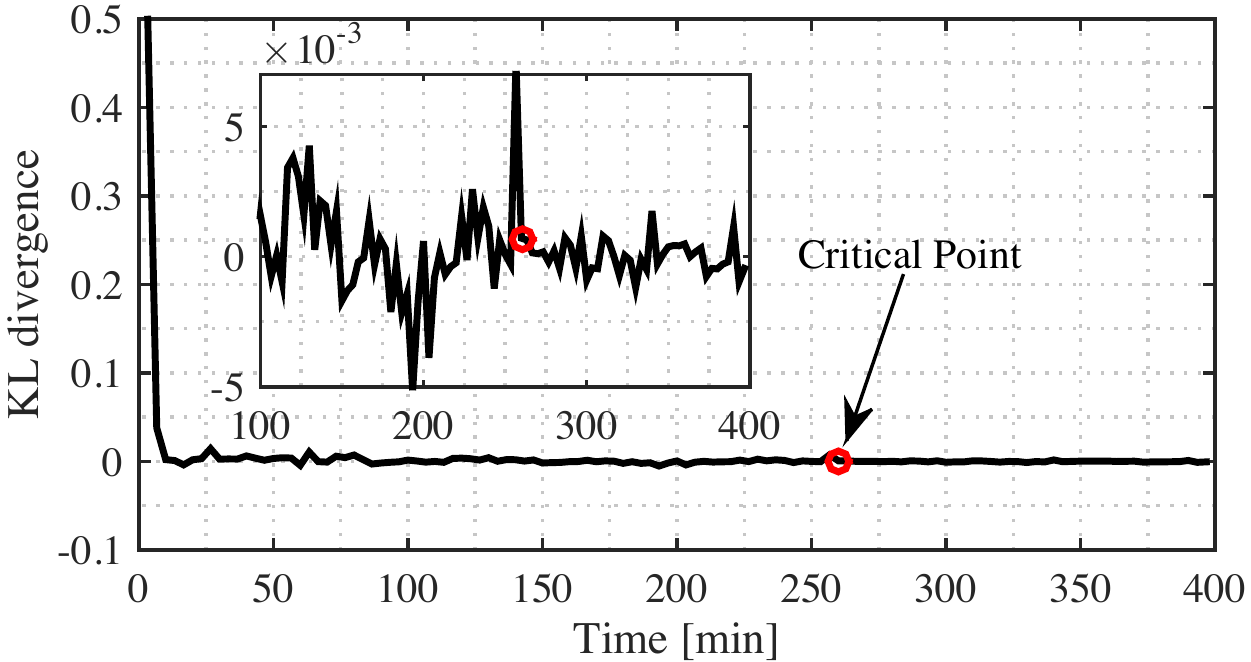}
		\caption{Relative speed, $ n^{\ast}_{\Delta v} \approx 1.56 \times 10^{5} $}
	\end{subfigure}
	~
	\begin{subfigure}[t]{0.48 \textwidth}
		\centering
		\includegraphics[scale = 0.66]{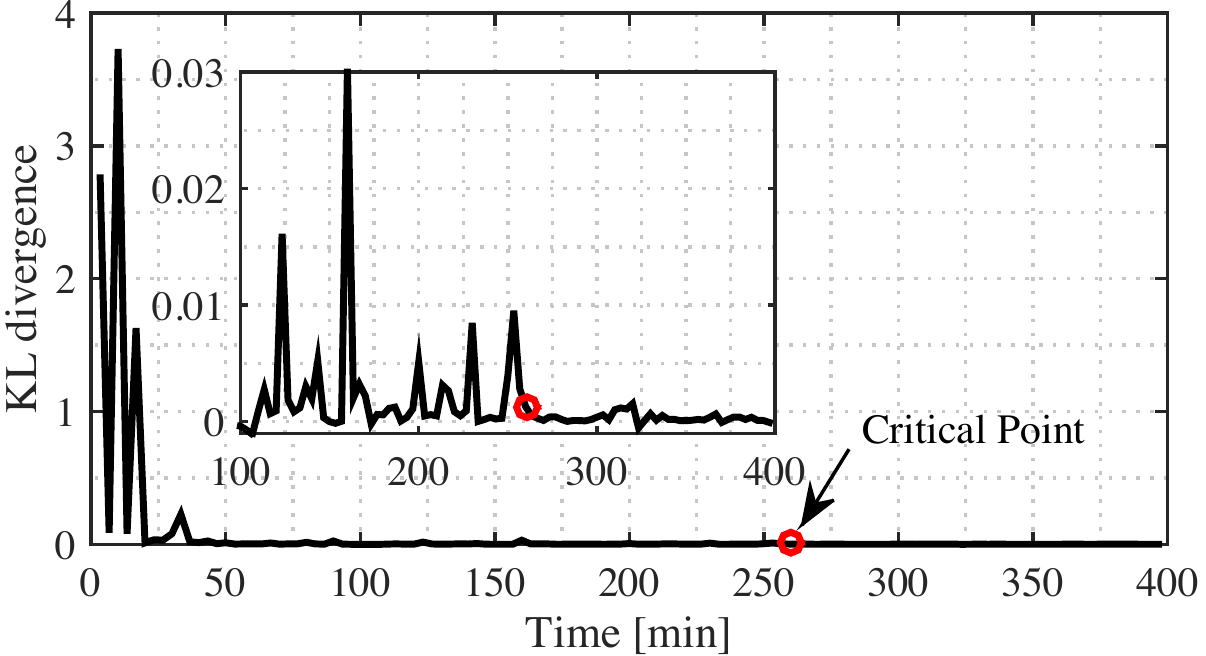}
		\caption{Speed, $ n^{\ast}_{v_e} \approx 5.8 \times 10^{4} $}
	\end{subfigure}
	~
	\begin{subfigure}[t]{0.48 \textwidth}
		\centering
		\includegraphics[scale = 0.66]{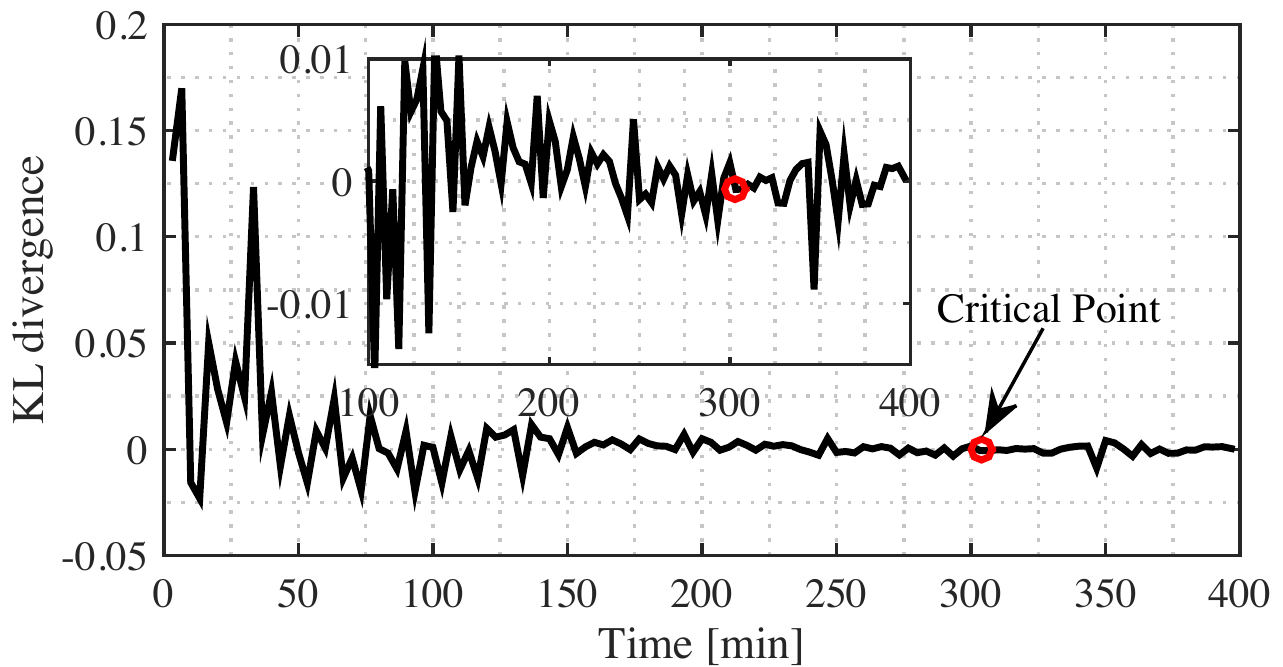}
		\caption{Acceleration, $ n^{\ast}_{a_{e}} \approx 1.82 \times 10^{5} $}
	\end{subfigure}
	\caption{The appropriate data amount of modeling driver's car-following behavior in terms of four variables for driver  \#12 with $ \epsilon = 10^{-4} $.}
	\label{fig:example_driver}
\end{figure*}

\begin{figure*}[t]
	\centering
	\begin{subfigure}{0.78\textwidth}
		\includegraphics[scale = 0.8]{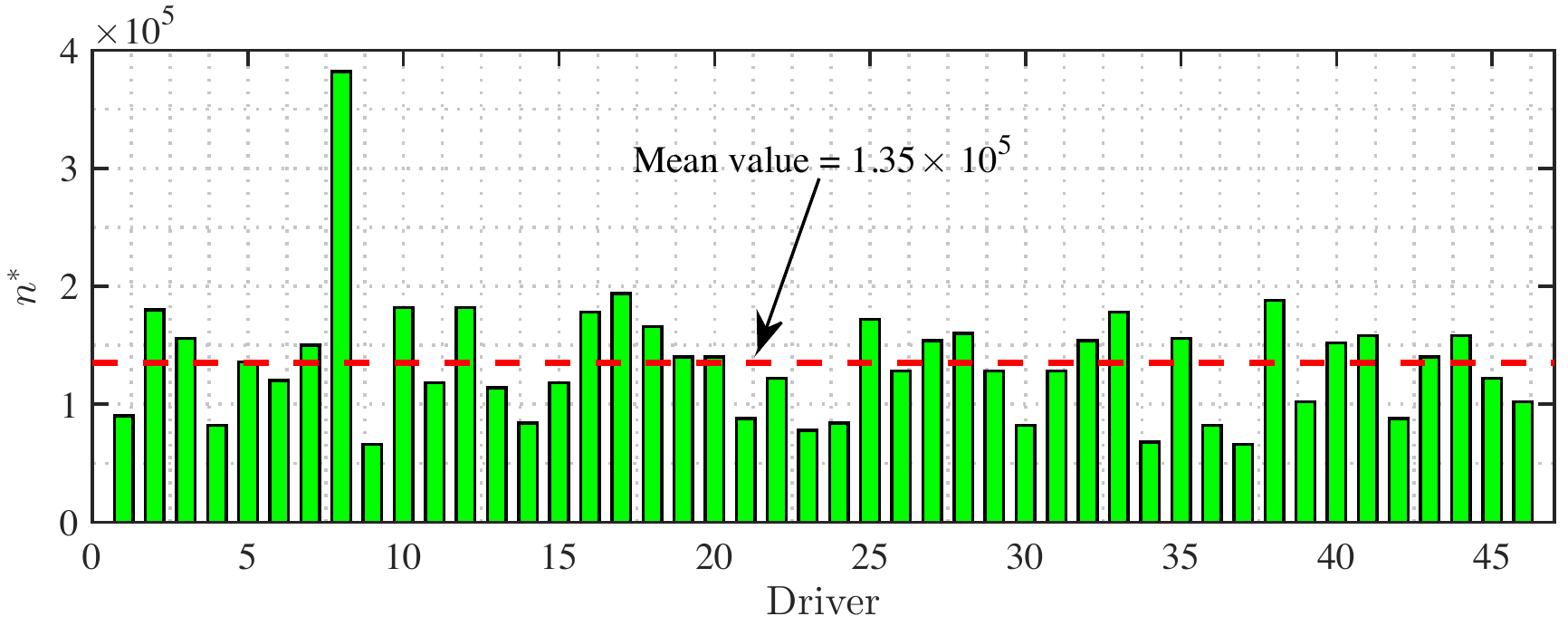}
		\caption{}
	\end{subfigure}
	\begin{subfigure}{0.19\textwidth}
		\includegraphics[scale = 0.8]{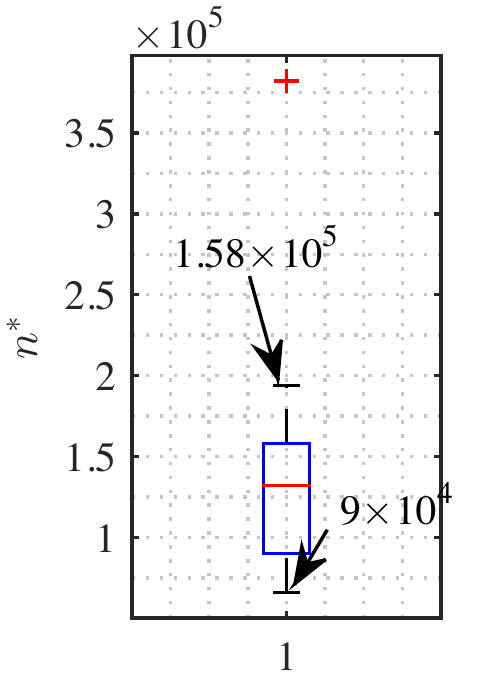}
		\caption{}
	\end{subfigure}
	\caption{The appropriate amount of NDD for all the participants in terms of modeling car-following behaviors with $ \epsilon = 10^{-4} $.}
	\label{fig: statistical_res}
\end{figure*}

\subsection{Data Processing}

Based on the definition and limitations of the car-following behavior, 46 drivers with 67,754 car-following events were extracted (Fig. \ref{fig:Data_Inf}). For modeling car-following behaviors, the variable selection varies by research topic. Different variable selection requires differing amounts of NDD. In this research, we apply the velocity $ v_{e} $ of the ego vehicle, the acceleration $ a_{e} $ of the ego vehicle, the relative speed $ \Delta v $, and the relative distance $ \Delta d $ between the ego vehicle and the preceding vehicle to formulate drivers' car-following behaviors, similar to \cite{Lefevre16}. For each variable, we compute the critical amount of driving data using (\ref{eq:KL_critical}).  To make the method more generalizable, we propose a max-minimum method to determine an appropriate amount of NDD. The appropriate amount of driving data that can fully cover driver behavior characteristics for each variable is computed by

\begin{equation}\label{eq:min}
{n}_{\{\star\}}^{\ast} = \min \{n|\mathrm{Equation}(\ref{eq:KL_critical}) \ \mathrm{is \ valid} \}
\end{equation}
with $ \{ \star\} \in \{ v_{e}, a_{e}, \Delta v, \Delta d \} $ and  $ m=2,000 $ in (\ref{eq:KL_critical}). According to (\ref{eq:KL_critical}) and (\ref{eq:min}), for each variable we can find an appropriate amount of NDD to cover the underlying characteristics. If researchers utilize a multivariate model to describe driver behaviors, the minimum amount of required NDD to cover driver behavior characteristics is the maximum value of all appropriate amount of these variables. Taking modeling the car-following behaviors for example, the appropriate amount of NDD using four variables can be computed by 

\begin{equation}\label{eq:max}
n^{\ast} = \max \{ n_{\{\star\}}^{\ast} | \{ \star\} \in \{ v_{e}, a_{e}, \Delta v, \Delta d \} \}
\end{equation}
Thus, we can obtain the optimal amount of NDD that can most effectively cover all the driving characteristics that we focus on by using the NDD as little data as possible. 

\subsection{Results Discussion and Analysis}

\subsubsection{Univariate Kernel Density Estimation}
Based on (\ref{eq:kernel_fun}), we obtain the kernel density for all variables with different amounts of data, as shown in Fig. \ref{fig:kernel_RelDis} -- Fig. \ref{fig:kernel_acc}. From the estimated results of kernel density with four variables, we note that when the amount of driving data is limited, the density changes greatly. For example, kernel densities greatly differ for relative distance, relative speed, speed and acceleration of the ego vehicle, when comparing $ n = 2,000 $ and $ n = 22,000 $, respectively. When the quantity of the data is larger, the divergences between densities with different data amounts are smaller. For example, the kernel densities with $ n = 82,000 $ and $ 102,000 $  are quite similar for every single variable. 
%
%

\subsubsection{Appropriate Amount of NDD}
To show the appropriate data amount of data for each variable, examples for driver \#12 are given for each single variable. The KL divergences for each variable are computed by (\ref{eq:KL}) and shown in Fig. \ref{fig:example_driver}. The red circle represents the critical value for each variable computed via (\ref{eq:KL_critical}). The vertical axis is the KL divergence value and the horizontal axis is the driving time, $ t $, of collecting data, computed by

\begin{equation}
t = \frac{n}{f\cdot 60}
\end{equation}
where $ n $ is the amount of data collected, $ f $ is the sample frequency, the unit of $ t $ is minute, and $ f = 10  $ Hz. We can conclude that the appropriate amounts of driving data with respect to $ \Delta d $, $ \Delta v $, $ v_e $ and $ a_e $ are $ 4.6 \times 10^{4}  (\approx 76.7\  \mathrm{min}) $, $ 1.56 \times 10^{5} (\approx 260 \ \mathrm{min}) $, $ 1.56 \times 10^{5} (\approx 260 \ \mathrm{min}) $, and $ 1.82\times 10^{5} (\approx 303.3 \ \mathrm{min}) $, respectively. Based on the results in Fig. \ref{fig:example_driver}, the appropriate amount of data for modeling the car-following behaviors of driver \#12 using four variables can be computed by (\ref{eq:max}) and obtained as $ n^{\ast}  = 1.82 \times 10^{5} $.

\begin{figure}[t]
	\centering
	\includegraphics[width = 0.45\textwidth]{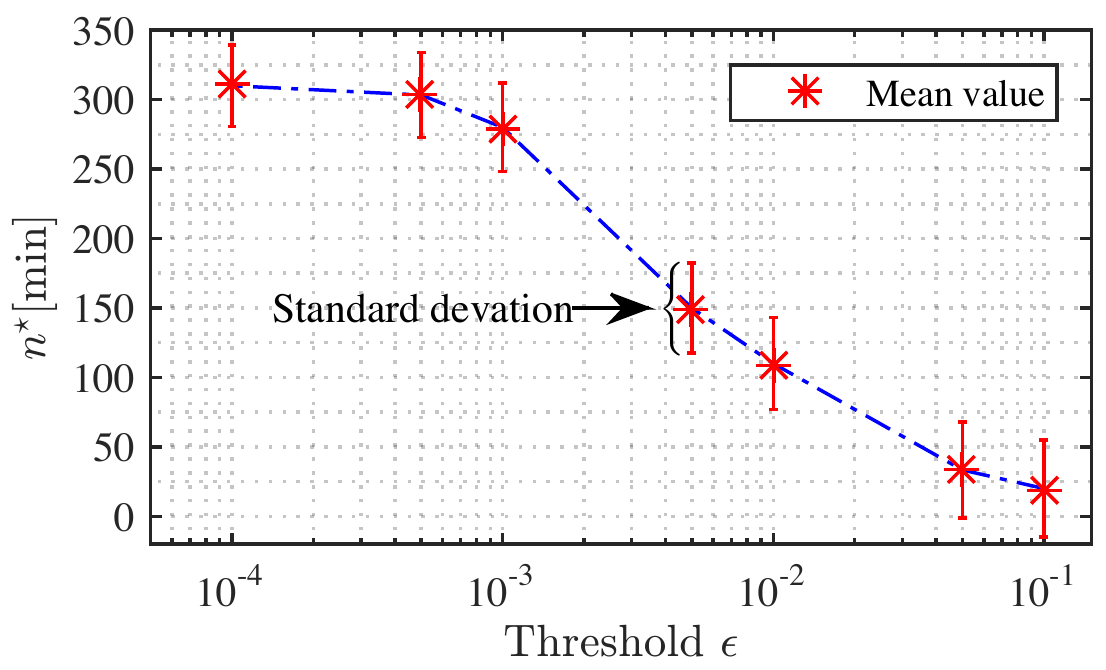}
	\caption{The statistical results of the influences of threshold $ \epsilon $ on data size for 46 drivers.}
	\label{Fig:Threshold}
\end{figure}

Fig. \ref{fig: statistical_res} shows the statistical results of the appropriate amount of NDD to model drivers' car-following behavior for all driver participants. We note that the appropriate amount of NDD to model the driver's car-following behavior using four variables is about $ 1.35\times 10^{5} $ ($ \approx  225.5 \ \mathrm{min} $). The suitable amount of NDD for modeling driver's car-following behavior ranges from $ 9.0\times 10^{4} (\approx  150 \ \mathrm{min})$ to $ 1.58\times 10^{5} (\approx  263.3 \ \mathrm{min})$, as shown in Fig. \ref{fig: statistical_res}(b). 

\subsubsection{Influence of Threshold $ \epsilon $ on Data Size}
According to (\ref{eq:KL_critical}) we know that the threshold $ \epsilon $ will affect the estimated data amount for understanding driver behavior. Fig. \ref{Fig:Threshold} presents the influences of threshold $ \epsilon $ on the estimated amount of NDD. We conclude that a larger threshold results in a smaller amount of NDD, and vice versa. When the threshold is less than $ 5\times 10^{-4} $, the amount of required NDD is convergent to a constant ($ \approx 300 $ min) for the car-following behaviors. Therefore, to obtain a conservative result, the threshold was set $ \epsilon < 10^{-3} $.
When $\epsilon = 5\times 10^{-4} $, the results ($ n^{\ast} \approx 300 $ min in Fig. \ref{fig: statistical_res} and Fig. \ref{Fig:Threshold}) from the methodology we propose in this paper are consistent with the results collected from the published papers ($ n^{\ast} \approx 288$ min in  Fig. \ref{TrAc_MDB}), which also support the claims based on our proposed methods.

\subsection{Multivariate KDE Method}

\begin{figure}[t]
	\centering
	\begin{subfigure}[t]{0.335\textwidth}
		\includegraphics[width = \textwidth]{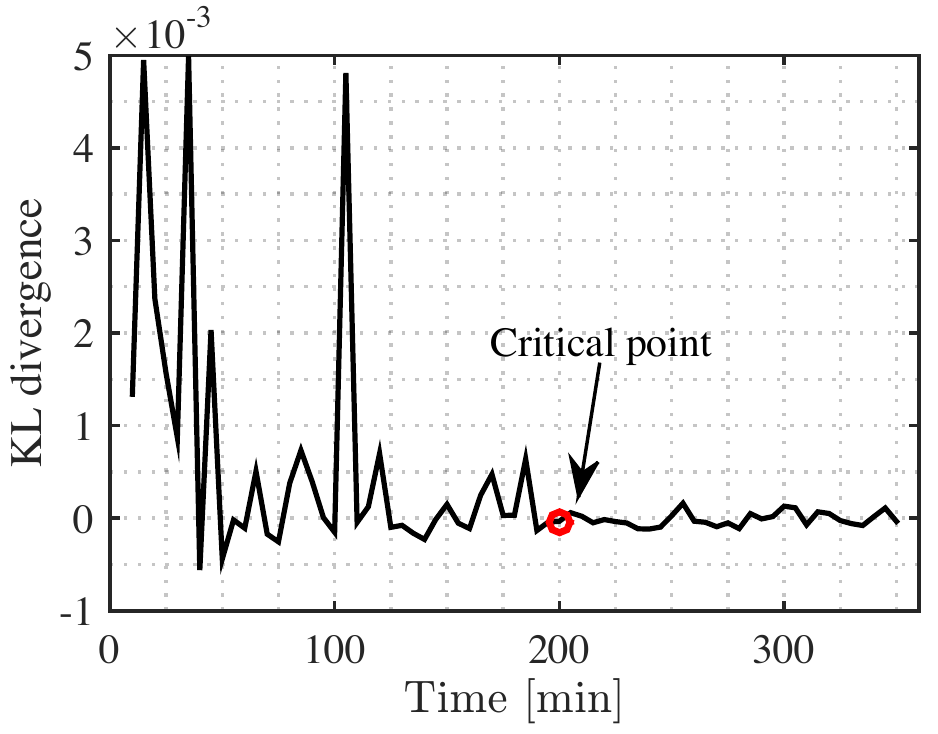}
	\end{subfigure}
	\begin{subfigure}[t]{0.145\textwidth}
		\includegraphics[width = \textwidth]{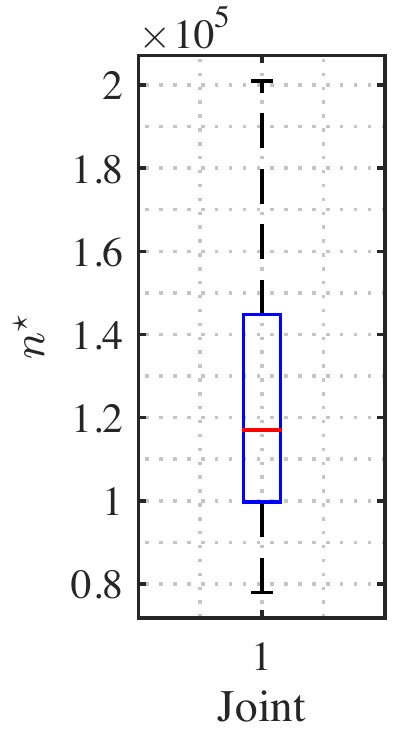}
	\end{subfigure}
	\caption{The KL divergence using the multivariate kernel density estimation method. Left: a case example; right: the statistical results for the critical point with $ \epsilon = 10^{-4} $.}
	\label{fig:MKDE}
\end{figure}

To support the proposed method, we also investigate the joint relationship between different variables using multivariate KDE \footnote[2]{This can be achieved by using Matlab command \texttt{mvksdensity}} method \cite{silverman1986density}. Thus, a multivariate kernel density, $\bm{\widehat{f}}(\bm{x}; n) $, with $ n $ amount of driving data is estimated, where $ \bm{x}\in \mathbb{R}^{4\times 1} $. To improve computing speed, we select 15 points for each variable as computing points, then obtaining $ N= 15^{4} $ vectors $ \{\bm{\widetilde{x}}_{i}\}_{i = 1}^{N} $ to compare the similarity between two multivariate kernel densities by

\begin{equation}\label{equ:MKDE}
\begin{split}
KL\left( \bm{\widehat{f}}(\bm{x}; n+m) || \bm{\widehat{f}}(\bm{x}; n)  \right)  = \\  \sum_{i = 1}^{N} \bm{\widehat{f}}(\bm{\widetilde{x}}_{i}; n+m) \log \frac{\bm{\widehat{f}}(\bm{\widetilde{x}}_{i}; n+m)}{\bm{\widehat{f}}(\bm{\widetilde{x}}_{i}; n)} \\
\end{split}
\end{equation}
Fig. \ref{fig:MKDE} demonstrates an example of the optimal amount of NDD that is enough to cover driver's car-following characteristics based on multivariate KDE and the statistical results of 21 drivers. We can know that the appropriate amount of driving data to model driver's car-following behavior using four variables is about $ n^{\ast} = 1.17\times 10^{5}$ ($ \approx $195 min). The right plot in Fig. \ref{fig:MKDE} demonstrates  that the suitable amount of NDD ranges from  $ 7.8 \times 10^4 $ ($ \approx 130 $ min) to $ 2.01\times 10^5 $ ($ \approx $ 335 min).

\begin{table}[t]
	\centering
	\caption{\textsc{The Optimal Amount of Demanded NDD Using Univariate and Multivariate KDE Methods.}}
	\begin{tabular}{c|ccc}
		\hline
		\hline
		& Median & Maximum & Minimum \\
		\hline
		Univariate KDE & 225.5 min &263.3 min &150.0 min \\
		Multivariate KDE & 195.0 min & 335.0 min & 130.0 min\\
		\hline
		\hline
	\end{tabular}
	\label{Table:CompSinglemulti}
\end{table}

Table \ref{Table:CompSinglemulti} compares the estimation results of the amount of required driving data for modeling car-following behavior using four variables based on univariate KDE and multivariate KDE. We note that the univariate KDE method and the multivariate KDE method obtain the appropriate data amount of 225.5 min and 195.0 min, respectively. The minimum amounts of required NDD using both methods are also similar (150.0 min and 130.0 min), but the univariate KDE method will slightly overestimate the required data amount, compared to the multivariate KDE method. 

However, the multivariate KDE method will exponentially increase the computation cost with increasing sampling data points of each variable. In the case with a four-dimension feature $ \bm{x} = [x_{1}, x_{2}, x_3, x_4]^{T} \in \mathbb{R}^{4\times 1} $, $ M $ sampling points of each variable are selected, i.e., $ \widetilde{x}_{i} = \{ \widetilde{x}_{i}^{1}, \cdots, \widetilde{x}_{i}^{M} \} $, where $ i = 1,2,3,4 $, then we will obtain $ M^{4} $ sampling feature vectors by meshing each variable to compute $ KL( \bm{\widehat{f}}(\bm{\widetilde{x}}; n+m) || \bm{\widehat{f}}(\bm{\widetilde{x}}; n)) $ in (\ref{equ:MKDE}). Compared to the multivariate KDE method, the univariate KDE method only requires  $ 4M $ sampling points in the same condition. For example, when $ M=100 $, the univariate KDE method only requires 400 data points, but the multivariate KDE method needs to compute $ 10^8 $ feature vectors. Therefore, in our case, the amount of sampling point in each variable is selected as $ 15 $ to compute the KL divergence when using the multivariate KDE method. A lower amount of sampling point in multivariate KDE method can shorten computing time but reduce the accuracy of estimating $ KL( \bm{\widehat{f}}(\bm{\widetilde{x}}; n+m) || \bm{\widehat{f}}(\bm{\widetilde{x}}; n)  ) $, which may result in no solutions for convergent condition (\ref{eq:KL_critical}). 


\section{Further Discussions}
In this paper, we point out and discuss the issues concerning the amount of data needed to understand and model driver behaviors, which is, to our best knowledge, the very first time to do so in literature. Question such as ``\textit{How much naturalistic driving data is sufficient for understanding and modeling driver behaviors?}'' is a basic issue that most researchers face. The methodology included in this paper can be used to assess the amount of data before modeling driver behaviors and designing a data-driven driving simulator. We provide a case study for the longitudinal driving behaviors to demonstrate the advantages of the proposed method. The approach could also be extended to the lateral driving behavior analysis such as lane change behavior. Other attributes are discussed below.

\subsection{Personalized Behavior}
In this paper, we focus primarily on modeling driver behaviors using the NDD collected from each single driver. We utilize the individual's driving data to model and understand individual driver behaviors that is also called personalized behaviors. The analysis and investigation based on all drivers' driving data for general driver behaviors were not involved in this paper. The methodology developed in this paper can also be directly applied to determining the requisite amount of data for establishing a general driver model, thus reducing the cost of experiments and resources. We will collect a broader range of driving data covering different ages, driving experience, and genders to investigate the difference in the  amount of required data for modeling between individual and general driver behavior.

\subsection{Small Probability Events}
The proposed assessment method for determining how much NDD is sufficient is feasible for modeling and understanding common driver behaviors such as car following, lane change, distractions/inattentions, or decision-making behaviors. But we have not investigated its application in research focusing on events at low probability, such as traffic accidents, because the small probability events has their own analysis approach \cite{kalra2016driving} differing from the proposed method in this paper.

\subsection{Feature Variable Selection}
As discussed in Section II, different formulation methods, including feature variable selection, lead to variety in the required amount of data. From (\ref{eq:max}), we know that the proposed method depends greatly on feature variable selection, which renders the proposed method more flexible. Let us take the car-following modeling of driver \#12 for example. When four variables are selected as shown in our case study, the appropriate amount of NDD is about 300 min; but when only three variables, e.g., relative distance, relative speed, and vehicle speed, are selected, the appropriate amount of NDD will be about 260 min (Fig. \ref{fig:example_driver}).

In this case study, we applied our approach to a limited number of scenarios. For example, stop-and-go scenarios were not included. However, we expect that the proposed methodology for determining how much data is enough to cover the features of driver behavior is relevant for a variety of scenarios, including stop-and-go.


%
%


\section{Conclusion}
In this paper, we focus on issues concerning the amount of data needed in naturalistic driving studies. To understand the diversity in the amount of data required for modeling driver behavior, we discuss and analyze the factors across different kinds of research. We propose a general method to determine the appropriate amount of driving data used for modeling driver behaviors from a statistical perspective. The Gaussian kernel density estimation approach is utilized and the Kullback-Liebler divergence method is employed to evaluate the similarity between two density functions with differing amounts of data. And then, a max-minimum method is applied to determine the appropriate amount of driving data. Last, a case study for modeling driver car-following behavior using the naturalistic driving data is conducted to demonstrate our proposed method. The proposed method in this paper and the conclusions from our experiment can provide researchers and engineers guidelines to design or conduct a naturalistic driving study.

However, thus far the proposed method does not suffice to reveal the correlated traffic dynamics over space and time. The method allows to determine the appropriate amount of driving data covering most of driving behaviors without considering correlated traffic dynamics and dynamic process in primitive behaviors. The development of a general method based on driving patterns and traffic dynamics to determine the amount of required driving data is our future work.




%
%

%


%
%

\ifCLASSOPTIONcaptionsoff
  \newpage
\fi



\bibliographystyle{IEEEtran}
\bibliography{bib_ITS.bib}
%
%

%

\begin{IEEEbiography}[{\includegraphics[width=1in,height=1.25in,clip,keepaspectratio]{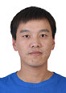}}]{Wenshuo Wang}
(S'15) received his B.S. in Transportation
Engineering from ShanDong University of
Technology, Shandong, China, in 2012. He is a
Ph.D. candidate for Mechanical Engineering, Beijing
Institute of Technology (BIT). Now he is a visiting
scholar studying in the School of Mechanical Engineering,
University of California at Berkeley (UCB).
His work focuses on modeling and recognizing drivers behavior, making
intelligent control systems between human driver and vehicle.
\end{IEEEbiography}

\begin{IEEEbiography}[{\includegraphics[width=1.2in,height=1.3in,clip,keepaspectratio]{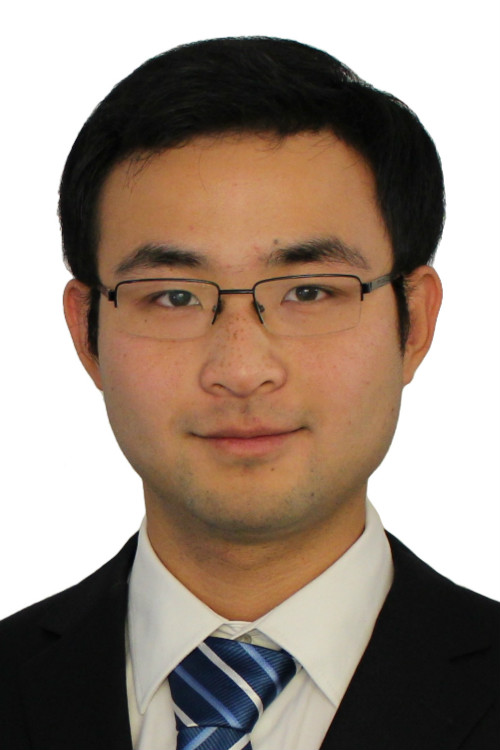}}]{Chang Liu} (S'15) received the B.S. degree
in Electrical Engineering and B.S. degree in Applied
Mathematics from Peking University, China,
in 2011. He received the M.S. degree in Mechanical
Engineering from the University of California at
Berkeley, CA, USA, in 2014, where he is also
currently working toward the Ph.D. degree in Mechanical
Engineering.
He is a Graduate Student Researcher with the Vehicle
Dynamics and Control Laboratory headed by
Prof. J Karl Hedrick. His research interests include
robot path planning, distributed estimation and human-robot collaboration.
\end{IEEEbiography}

\begin{IEEEbiography}[{\includegraphics[width=1in,height=1.25in,clip,keepaspectratio]{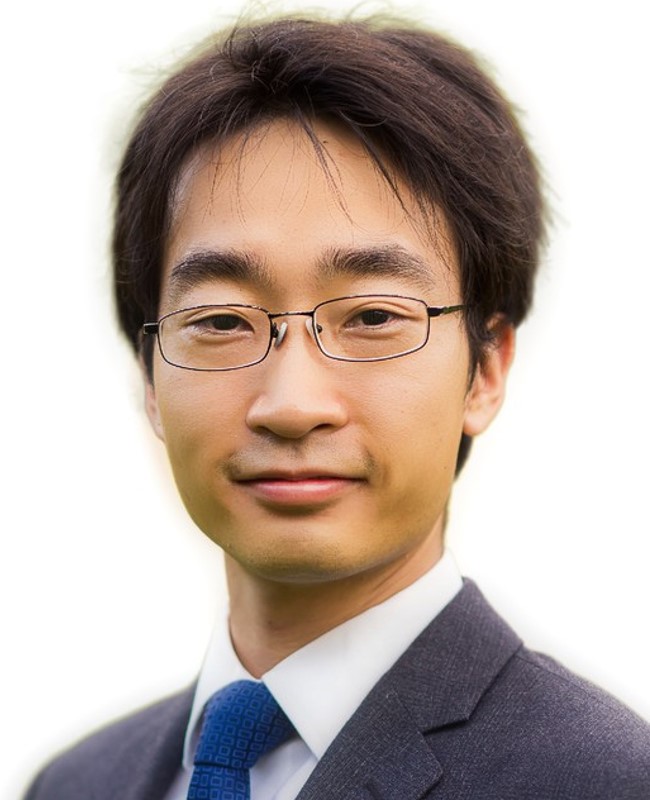}}]{Ding Zhao} received his Ph.D. degree in 2016 from the University of Michigan, Ann Arbor. He is currently an Assistant Research Scientist at Mechanical Engineering of the University of Michigan. His research interests include the autonomous vehicles, intelligent transportation, connected vehicles, dynamics and control, human-machine interaction, machine learning, and big data analysis.
\end{IEEEbiography}




\end{document}